\pdfoutput=1

\documentclass[sigconf]{acmart}
\AtBeginDocument{%
  }

\setcopyright{acmlicensed}
\copyrightyear{YEAR}
\acmYear{YEAR}
\acmConference[Conference acronym 'XX]{Make sure to enter the correct conference title from your rights confirmation emai}{MONTH XX--XX, YEAR}{LOCATION, XX}
\acmISBN{978-1-4503-XXXX-X/18/06}

\usepackage[textsize=small]{todonotes}
\usepackage{algorithm}
\usepackage{algorithmic}

\usepackage{amsmath}
\usepackage{amsfonts}
\usepackage{algorithmic}
\usepackage{graphicx}
\usepackage{subcaption}
\usepackage{textcomp}
\usepackage{xcolor}
\usepackage{hyperref}
\usepackage{multirow}
\usepackage{tabularx}                     
\newcolumntype{L}[1]{>{\raggedright\arraybackslash}p{#1}}
\usepackage{placeins}
\usepackage{stfloats}      

\newcommand{\Prob}{\mathbb{P}}
\newcommand{\E}{\mathbb{E}}
\newcommand{\R}{\mathbb{R}}
\newcommand{\X}{\mathcal{X}}
\newcommand{\Y}{\mathcal{Y}}
\newcommand{\paramspace}{\boldsymbol{\Theta}}
\newcommand{\params}{\boldsymbol{\theta}}

\newcommand{\model}{f_{\params}}
\newcommand{\cfmodel}{g_{\params}}
\newcommand{\inst}{\boldsymbol{x}}
\newcommand{\cfinst}{\widetilde{\boldsymbol{x}}}

\newcommand{\dataset}{\mathcal{D}}

\newcommand{\Loss}{\mathcal{L}}

\newcommand{\loss}{\ell}

\newcommand{\argmin}{\text{arg\,min}}

\newcommand{\perturb}{\boldsymbol{\delta}}

\begin{document}

\title[Countering Overfitting with Counterfactual Examples]{Countering Overfitting with Counterfactual Examples}

\author{Flavio Giorgi}
\email{giorgi@di.uniroma1.it}
\affiliation{%
  \institution{Sapienza University of Rome}
  \city{Rome}
  \country{Italy}
}
\orcid{0009-0003-6654-4879}
\author{Fabiano Veglianti}
\email{fabiano.veglianti@uniroma1.it}
\affiliation{%
  \institution{Sapienza University of Rome}
  \city{Rome}
  \country{Italy}
}
\orcid{0009-0007-1563-4953}

\author{Fabrizio Silvestri}
\email{fsilvestri@diag.uniroma1.it}
\affiliation{%
  \institution{Sapienza University of Rome}
  \city{Rome}
  \country{Italy}
}
\orcid{0000-0001-7669-9055}

\author{Gabriele Tolomei}
\email{tolomei@di.uniroma1.it}
\affiliation{%
  \institution{Sapienza University of Rome}
  \city{Rome}
  \country{Italy}
}
\orcid{0000-0001-7471-6659}

\renewcommand{\shortauthors}{Giorgi et al.}

\begin{abstract}
Overfitting is a well-known issue in machine learning that occurs when a model struggles to generalize its predictions to new, unseen data beyond the scope of its training set. 
Traditional techniques to mitigate overfitting include early stopping, data augmentation, and regularization.
In this work, we demonstrate that the degree of overfitting of a trained model is correlated with the ability to generate \textit{counterfactual examples}. The higher the overfitting, the easier it will be to find a valid counterfactual example for a randomly chosen input data point.
Therefore, we introduce CF-Reg, a novel regularization term in the training loss that controls overfitting by ensuring enough margin between each instance and its corresponding counterfactual.
Experiments conducted across multiple datasets and models show that our \textit{counterfactual regularizer} generally outperforms existing regularization techniques.
\end{abstract}




\maketitle

\section{Introduction}
\label{sec:intro}
One of the key challenges in machine learning is developing models that can generalize their predictions to new, unseen data beyond the scope of the training set. 
When the predictive accuracy of a model on the training set far exceeds that on the test set, this indicates a phenomenon known as \textit{overfitting}.
In general, the impact of overfitting is more pronounced for highly complex models like recent deep neural networks with billions of parameters. 
To compensate for the risk of overfitting, these models require massive amounts of training data, which may not always be feasible. 

Therefore, several strategies have been proposed in the literature to mitigate the problem of model overfitting. Among these, \textit{early stopping} interrupts the training phase before the model starts learning the noise in the data rather than the actual underlying input/output relationship.
Furthermore, \textit{data augmentation} is a technique that artificially increases the training set. For example, in the context of image data, this can include applying translation, flipping, and rotation transformations to input samples.
Finally, \textit{regularization} is a collection of training/optimization techniques that try to eliminate irrelevant factors by assessing the importance of features, preventing minor input changes from causing significant output variations.

In this work, we offer an entirely new perspective on model overfitting, establishing a connection with the ability to generate \textit{counterfactual examples}~\cite{wachter2017hjlt}. The notion of counterfactual examples has been successfully used, for instance, to attach post-hoc explanations for predictions of individual instances in the form: ``\textit{If A had been different, B would \textbf{not} have occurred}''~\citep{stepin2021survey}. 
Generally, finding the counterfactual example for an instance resorts to searching for the minimal perturbation of the input that crosses the decision boundary induced by a trained model. This task reduces to solving a constrained optimization problem.

\begin{figure}
\centering
\begin{subfigure}[htb!]{.4\linewidth}
\includegraphics[width=.9\linewidth]{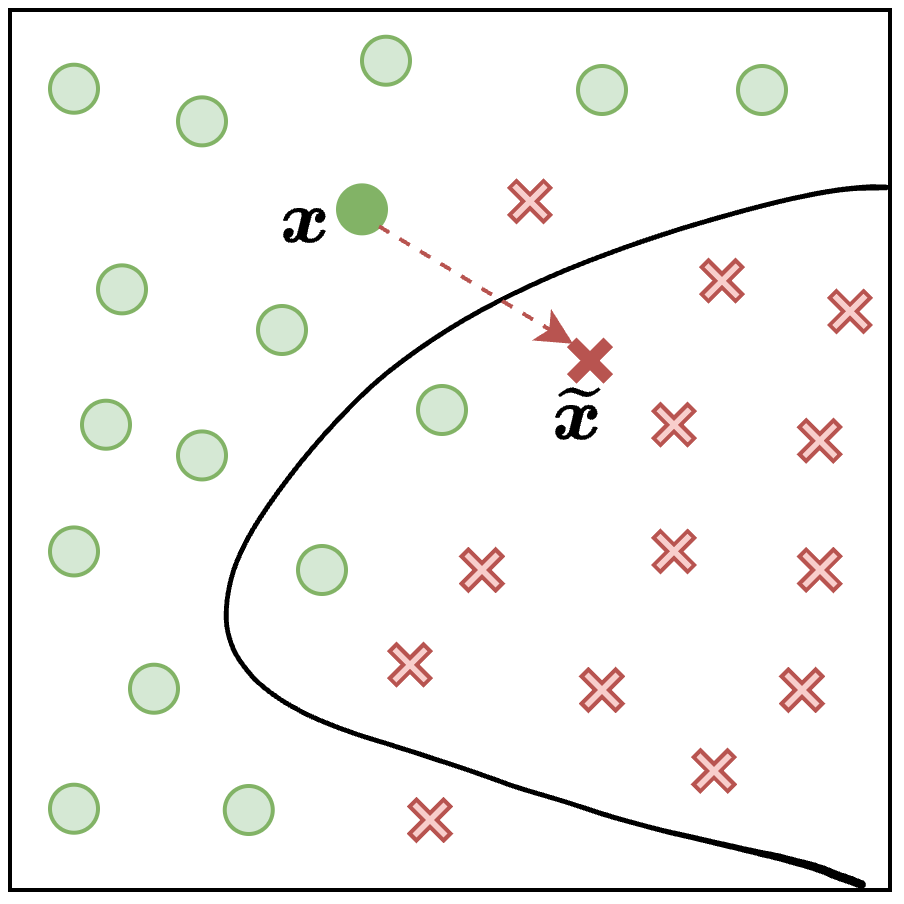}
\caption{No overfitting}
\end{subfigure}
\qquad
\begin{subfigure}[h]{.4\linewidth}
\includegraphics[width=.9\linewidth]{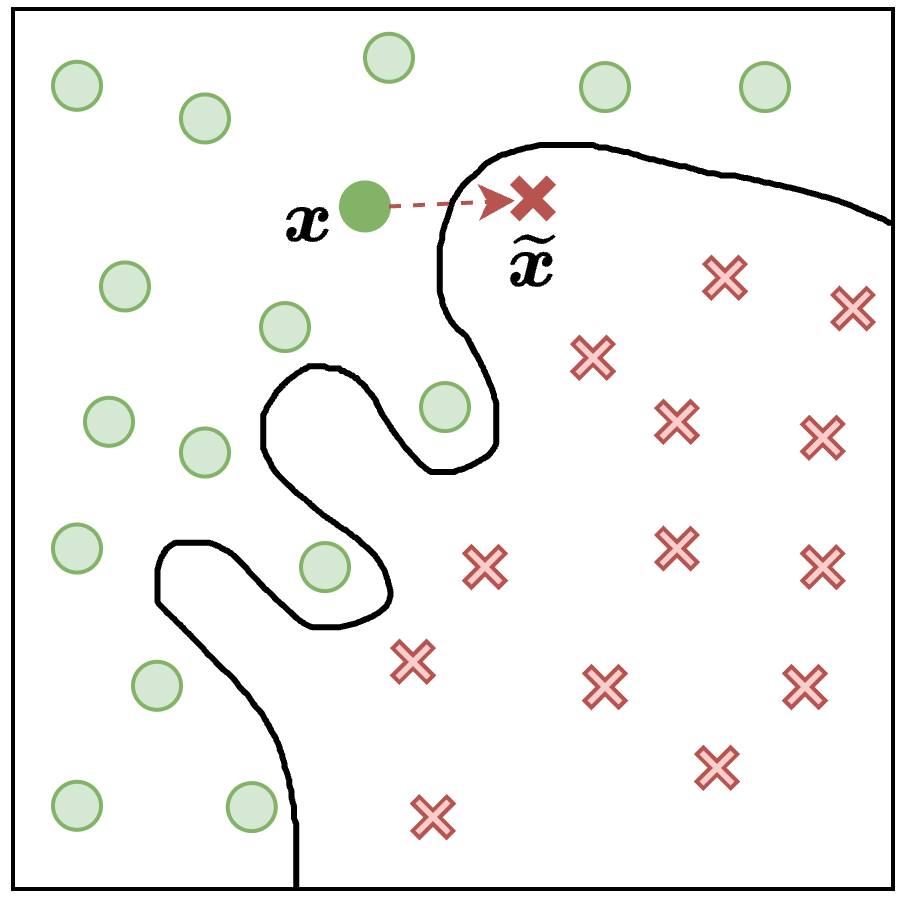}
\caption{Overfitting}
\end{subfigure}%
\caption{Distance between an input data point ($\inst$) and its counterfactual example ($\cfinst$): On average, this may be higher for a well-trained model (a) than an overfitted model (b).}
\label{fig:intuition}
\end{figure}

In the presence of a strong degree of model overfitting, the decision boundary learned becomes a highly convoluted surface, up to the point where it perfectly separates every training input sample. 
Therefore, each data point, on average, is ``closer'' to the decision boundary, making it easier to find the best counterfactual example. 
The intuition behind this claim is depicted in Figure~\ref{fig:intuition} and grounded in the theoretical foundations of margin theory \cite{wu2019arxiv}.

Following this idea, we introduce a novel \textit{counterfactual regularization} term in the training loss that controls overfitting by enforcing a margin between each instance and its hypothetical counterfactual.
Below, we summarize the primary novel contributions of our work:
\begin{itemize}
\item[$(i)$] We explore the relationship between generalizability and counterfactual explanations, based on margin theory.
\item[$(ii)$] We show that counterfactual examples can effectively guide and enhance the model training process.
\item[$(iii)$] We propose a counterfactual regularizer (CF-Reg) that outperforms existing regularization techniques.
\item[$(iv)$] We cast our regularization method in an extensible framework that is compatible with any differentiable counterfactual example generator. The source code for our method is available at the following GitHub repository: \url{https://anonymous.4open.science/r/CF-Reg-5AB6/README.md}.

\end{itemize}

The remainder of this paper is structured as follows. In Section~\ref{sec:related}, we summarize related work. Section~\ref{sec:background} reviews background concepts, useful for our problem formulation in Section~\ref{sec:problem}. In Section~\ref{sec:method}, we describe our method that is validated through extensive experiments in Section~\ref{sec:experiments}. 
We discuss the applicability and limitations of our method in Section~\ref{sec:discussion}.
Finally, Section~\ref{sec:conclusion} concludes our work.

\section{Related Work}
\label{sec:related}

\subsection{Model Overfitting}
\label{subsec:overfitting}
%
%
Several strategies have been proposed in the literature to reduce the effects of overfitting, which can be broadly categorized as follows.

\smallskip
\noindent \textit{\textbf{Early Stopping.}} This strategy aims to mitigate the phenomenon known as ``learning speed slow-down.'' This issue occurs when the accuracy of a model ceases to improve or even worsens due to noise-learning. The concept has a long history, dating back to the 1970s in the context of the Landweber iteration~\citep{raskutti2011ccc}. It has since been widely adopted in iterative algorithms, particularly for training deep neural networks in combination with backpropagation~\citep{caruana2000neurips}.

\smallskip
\noindent \textit{\textbf{Data Augmentation.}} The more complex the model, the higher the number of parameters that need to be learned. Therefore, the size of the training set must be adequate to avoid overfitting. Data augmentation techniques play a crucial role in enhancing model generalization across various domains, including pattern recognition and image processing. These techniques aim to expand existing datasets to generate additional data. Typically, four main approaches are employed~\citep{sun2014cvpr,karystinos2000tnn,yip2008bio}: \textit{(i)} acquiring new data, \textit{(ii)} introducing random noise to the existing dataset, \textit{(iii)} reprocessing existing data to produce new instances, and \textit{(iv)} sampling new data based on the distribution of the existing dataset.

\smallskip
\noindent \textit{\textbf{Regularization.}} An overfitting model tends to incorporate all available features into its decision-making process, even those with minimal impact or that are simply noise. To address this issue, there are two main approaches: \textit{(i)} \textit{Feature selection}, where only the most relevant features are retained, discarding those deemed irrelevant; \textit{(ii)} \textit{Feature regularization}, which involves minimizing the influence of less important features by reducing their weights in the model.
However, since identifying useless features can be challenging, regularization techniques work by applying a penalty term, or ``regularizer'', to the objective function. This penalizes complex models, encouraging simpler and more generalizable solutions. For example, L1 (``Lasso'') and L2 (``Ridge'') regularization, which add the $L_1$-norm and $L_2$-norm of the learned parameter as a penalties, respectively, are commonly adopted in linear regression. Furthermore, \textit{Dropout} is a popular technique to combat overfitting in neural networks. This approach randomly drops units and relevant connections from the neural network during training to prevent co-adaptation~\citep{warde-farley2014iclr}.

Unlike existing regularization techniques, which either aim to reduce the magnitude of the learned model's weights (such as Lasso and Ridge) or randomly deactivate certain parameters (such as Dropout), our penalty term is inherently \textit{data-driven}. In other words, the counterfactual regularization we introduce (CF-Reg) is computed directly from the training observations, making it highly tailored to the specific dataset under consideration.

\smallskip
\noindent \textit{\textbf{Adversarial Training.}} Adversarial training is another strategy that can be adapted to mitigate overfitting. Notably, \citeauthor{madry2018pgd} \cite{madry2018pgd} proposed a min-max optimization framework that enhances model robustness by training against adversarial examples generated via Projected Gradient Descent (PGD). This approach serves as an effective form of regularization, as it discourages the learned model from depending on non-robust or spurious patterns \citep{stutz2019disentangling}.




\subsection{Counterfactual Examples}
\label{subsec:cf-examples}
Counterfactual examples have gained significant attention in machine learning due to their utility in various applications, including model \textit{explainability} and \textit{robustness}. 

\smallskip
\noindent \textit{\textbf{Explainability.}} Counterfactual explanations offer valuable insights into model predictions by providing alternative scenarios under which the prediction would change~\citep{wachter2017hjlt}. 
Several studies have explored the use of counterfactual examples to enhance the interpretability of complex machine learning models, such as ensembles of decision trees~\citep{tolomei2017kdd,tolomei2021tkde,lucic2022aaai} and deep neural networks (DNNs)~\citep{le2020kdd}, including graph neural networks (GNNs)~\citep{lucic2022aistats,chen2025tors}. Counterfactual instances may elucidate the underlying decision-making process of black-box models, enhancing trust and transparency~\citep{guidotti2018arxiv,karimi2020pmlr,chen2022cikm,movin2025tai}.


\smallskip
\noindent \textit{\textbf{Robustness.}} Counterfactual examples have also been leveraged to enhance the robustness of machine learning models against adversarial attacks~\citep{brown2018arxiv}. Indeed, there is a strict relationship between adversarial and counterfactual examples~\citep{freiesleben2022mm}, although their primary goals are divergent. While both adversarial and counterfactual examples involve perturbing input data to influence model predictions, adversarial examples are crafted to jeopardize the model, whereas counterfactual examples are generated to understand the model's behavior.
By generating instances that are semantically similar to the original input but induce different model predictions, \citet{he2019acsac} enhance model robustness against adversarial perturbations. 


Recent work has explored the use of generative models to produce realistic counterfactual instances for data augmentation and model refinement~\citep{ganin2016jmlr,hjelm2019iclr}, highlighting the potential of counterfactual reasoning in generative modeling tasks~\citep{lopez2020msb}.

To the best of our knowledge, however, this is the first study to link counterfactual examples to model generalization and to employ them as the cornerstone of a new regularization strategy.

\section{Background and Preliminaries}
\label{sec:background}
Let $\model: \X \mapsto \Y$ denote a predictive model parameterized by a set of (learnable) weights $\params \in \paramspace$ that takes an input $\inst\in \X\subseteq \R^n$ and maps it to an output $y\in \Y$. 
In the standard supervised learning setting, the optimal weights ($\params^*$) are found by minimizing a specific loss function ($\Loss$) computed on a training set ($\dataset$) of $m$ i.i.d. labeled instances, $\dataset=\{(\inst_i, y_i)\}_{i=1}^m$.
\begin{equation}
\label{eq:train}
\params^* = \argmin_{\params}\Big\{\Loss(\params;\dataset)\Big\} = \argmin_{\params}\Bigg\{\frac{1}{|\dataset|} \sum_{i=1}^m\loss(f_{\params}(\inst_i),y_i)\Bigg\}.
\end{equation}
Here, $\loss$ represents an instance-level error between the model's prediction for a given input ($\inst_i$) and its corresponding actual label ($y_i$), such as cross-entropy (for classification tasks) or mean squared error (for regression tasks).

Furthermore, we assume to have available a counterfactual generator model $\cfmodel: \X \mapsto \X$, for the predictive model $\model$, that takes as input a data point $\inst$ and produces as output its corresponding (optimal) counterfactual $\cfinst^*$. 
This typically requires solving a constrained objective as follows:


\begin{equation}
\label{eq:cf-train}
\begin{aligned}
\cfmodel(\inst) = \cfinst^* = \argmin_{\cfinst\in \X}  &\delta(\inst, \cfinst)\\
\text{ s.t.: } &\model(\inst) \neq \model(\cfinst).
\end{aligned}
\end{equation}

Here, $\delta:\X\times \X \mapsto \R_{\geq 0}$ is a function that measures the distance between the original input data point and its counterfactual. In practice, $\delta$ often computes the $L_1$- or $L_2$-norm of the displacement vector between the original and the counterfactual example.


In general, for a given input $\inst$, there could be several, possibly infinitely many \textit{valid} counterfactuals: $\cfinst_1, \cfinst_2, \ldots$ In this work, a valid counterfactual is considered as any instance $\cfinst$ that crosses the decision boundary induced by $\model$, i.e., where $\model(\inst) \neq \model(\cfinst)$. More refined notions of validity are also conceivable, such as those entailing the \textit{plausibility} of the generated counterfactual, which assesses whether the counterfactual example is indeed realistic. For example, suppose we have trained an image classifier to distinguish between birds and rabbits. In this context, a plausible (and valid) counterfactual for a rabbit must be an image resembling a bird.
However, for the purpose of this work, we are only interested in characterizing the ability to find a valid counterfactual and relating this to the degree of model overfitting, regardless of its plausibility.

Formally, let $\inst \in \X$ be an input sample, $\model$ a trained model, $\delta:\X\times \X \mapsto \R_{\geq 0}$ a distance function, and $\varepsilon \in \R_{>0}$ a fixed threshold. We consider an $\varepsilon$-\textit{valid counterfactual example} ($\varepsilon$-VCE) for $\inst$  any $\cfinst\neq \inst$, such that $\model(\inst) \neq \model(\cfinst)$ and $\delta(\inst, \cfinst) \leq \varepsilon$.

\section{Impact of Counterfactual Examples on Model Generalizability}
\label{sec:problem}

\subsection{Intuitive Perspective}
We consider a training set $\dataset = \{(\inst_i, y_i)\}_{i=1}^m$ of $m$ i.i.d. labeled instances, as introduced above. Then, suppose this dataset is used to train a sequence of $k$ different models $(f_{\params_0}, f_{\params_1}, \ldots, f_{\params_{k-1}})$. Each model $f_{\params_i}$ is associated with a training accuracy $\alpha_i$ (calculated on $\dataset$), such that $\alpha_0 < \alpha_1 < \ldots < \alpha_{k-1}$, where $f_{\params_0}$ indicates the random baseline model and $f_{\params_{k-1}}$ is the dummy model that simply memorizes the entire training set and, therefore, achieves the perfect training accuracy ($\alpha_{k-1} = 1$).\footnote{In practice, the sequence of $k$ models may correspond to intermediate checkpoints saved at various training epochs.}

We claim that for a fixed positive threshold $\varepsilon > 0$, the expected fraction of training points for which we can find an $\varepsilon$-valid counterfactual example is positively correlated with the training accuracy of the model. 
In other words, given two trained models $f_{\params_i}$ and $f_{\params_j}$, where $i > j$, whose training accuracies are $\alpha_i > \alpha_j$, we expect to find, on average, more $\varepsilon$-valid counterfactual examples for $f_{\params_i}$.

This intuition stems from the idea that a model with a higher training accuracy tends to have a more intricate decision boundary surface, which captures all the nuances of the training data more precisely. Consequently, data points are, \textit{on average}, closer to such a convoluted decision boundary, making it easier to cross it and thus to find valid counterfactual examples within a distance of $\varepsilon$.

In essence, a model for which finding counterfactual examples is ``too easy'' may indicate a risk of overfitting. Thus, a trade-off must exist between the model's generalizability and its counterfactual explainability, which we aim to investigate further in this study.

\subsection{Theoretical Motivation}
\label{subsec:theory}
The intuition outlined above is supported by a well-established theoretical foundation rooted in classical margin theory. Statistical learning theory tells us that a model's generalization error can be bounded using empirical margin distributions \cite{koltchinskii2002AoS}. A prominent example is the support vector machine (SVM), which \textit{explicitly} maximizes the minimum margin on linearly separable datasets.

Interestingly, margin maximization is not exclusive to SVMs. \citeauthor{soudry2018jmlr}~\cite{soudry2018jmlr} show that gradient descent applied to logistic regression on linearly separable data \textit{implicitly} maximizes the minimum margin, thereby mirroring the behavior of SVMs. This implicit bias toward margin maximization has also been observed in more complex models, including deep neural networks \cite{gunasekar2018neurips, ziwei2019iclr}.

Building on these insights, \citeauthor{wu2019arxiv}~\cite{wu2019arxiv} uncover an intriguing trade-off between the minimum and \textit{average} margins, where the latter is defined as the average distance of training examples to the decision boundary. Specifically, they show that optimizing for a larger minimum margin can unintentionally reduce the average margin, thereby increasing the model's vulnerability to adversarial examples. 
For the same reason, as training progresses and the average distance between data points and the decision boundary decreases, it becomes easier to identify a valid counterfactual example within a fixed perturbation radius $\varepsilon$. 

\begin{figure*}[h!]
    \centering 

    \begin{subfigure}{0.20\linewidth}
        \centering
        \includegraphics[width=\linewidth]{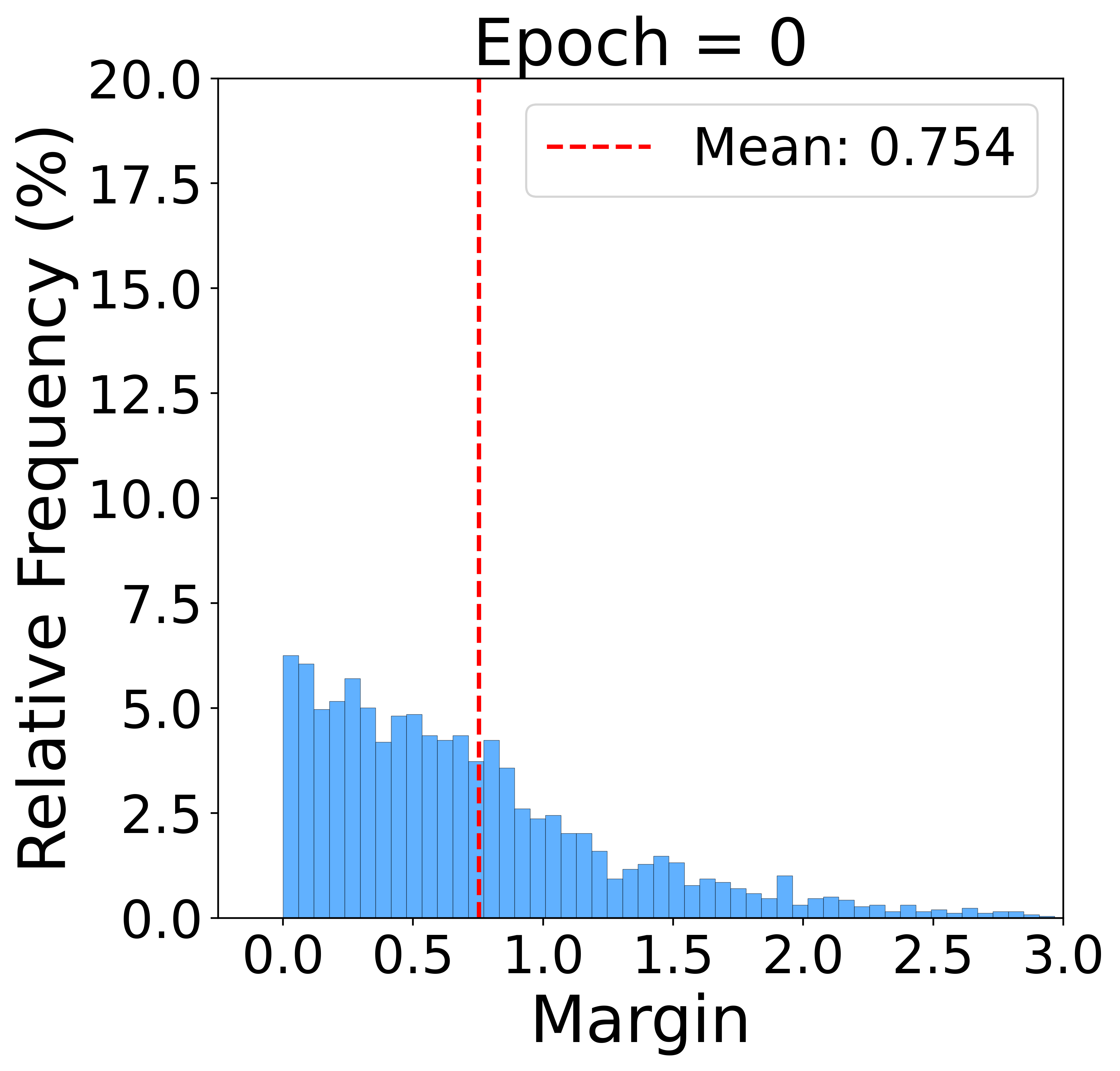}
    \end{subfigure}%
    \begin{subfigure}{0.19\linewidth}
        \centering
        \includegraphics[width=\linewidth]{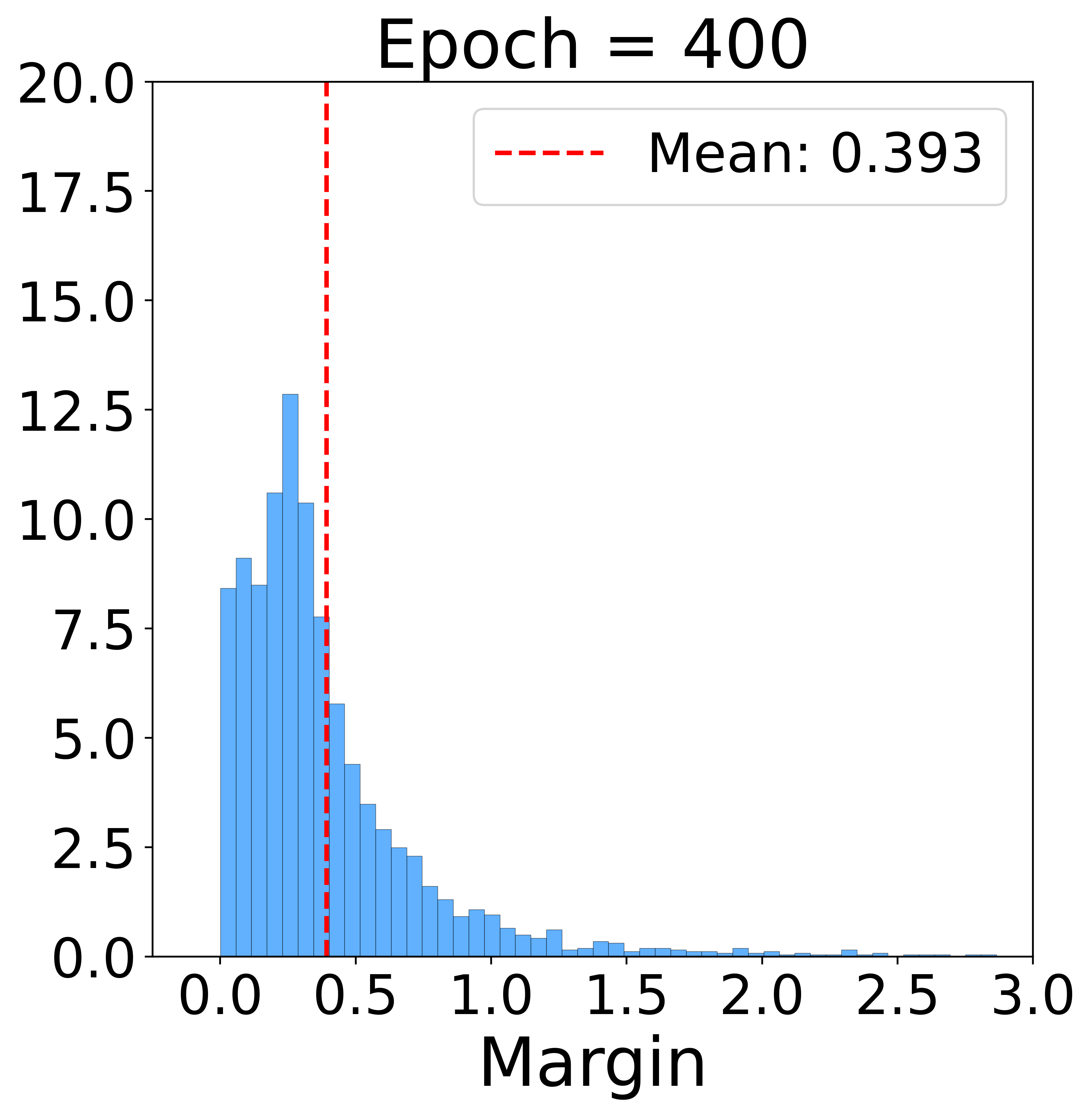}
    \end{subfigure}%
    \begin{subfigure}{0.19\linewidth}
        \centering
        \includegraphics[width=\linewidth]{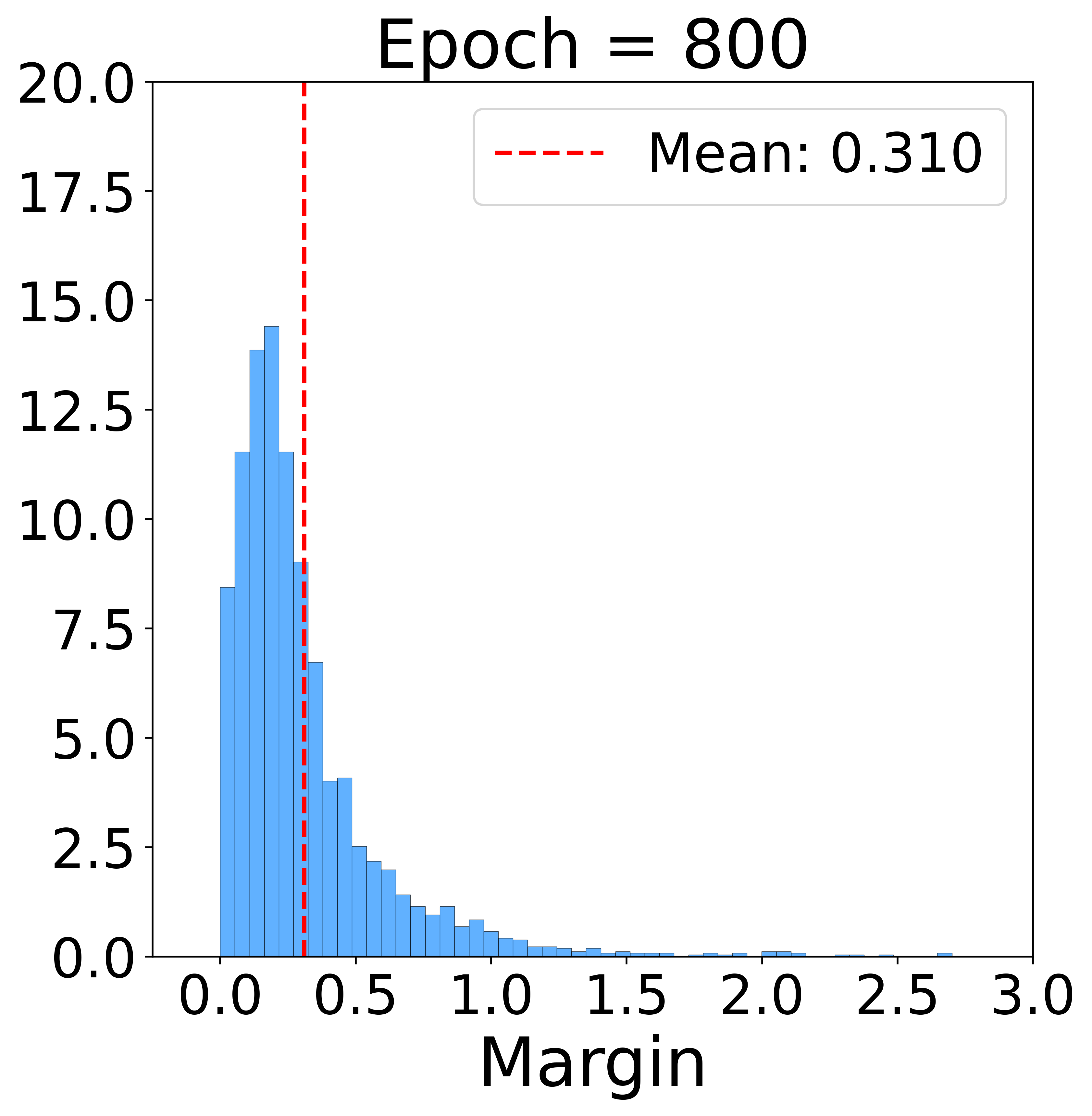}
    \end{subfigure}%
    \begin{subfigure}{0.19\linewidth}
        \centering
        \includegraphics[width=\linewidth]{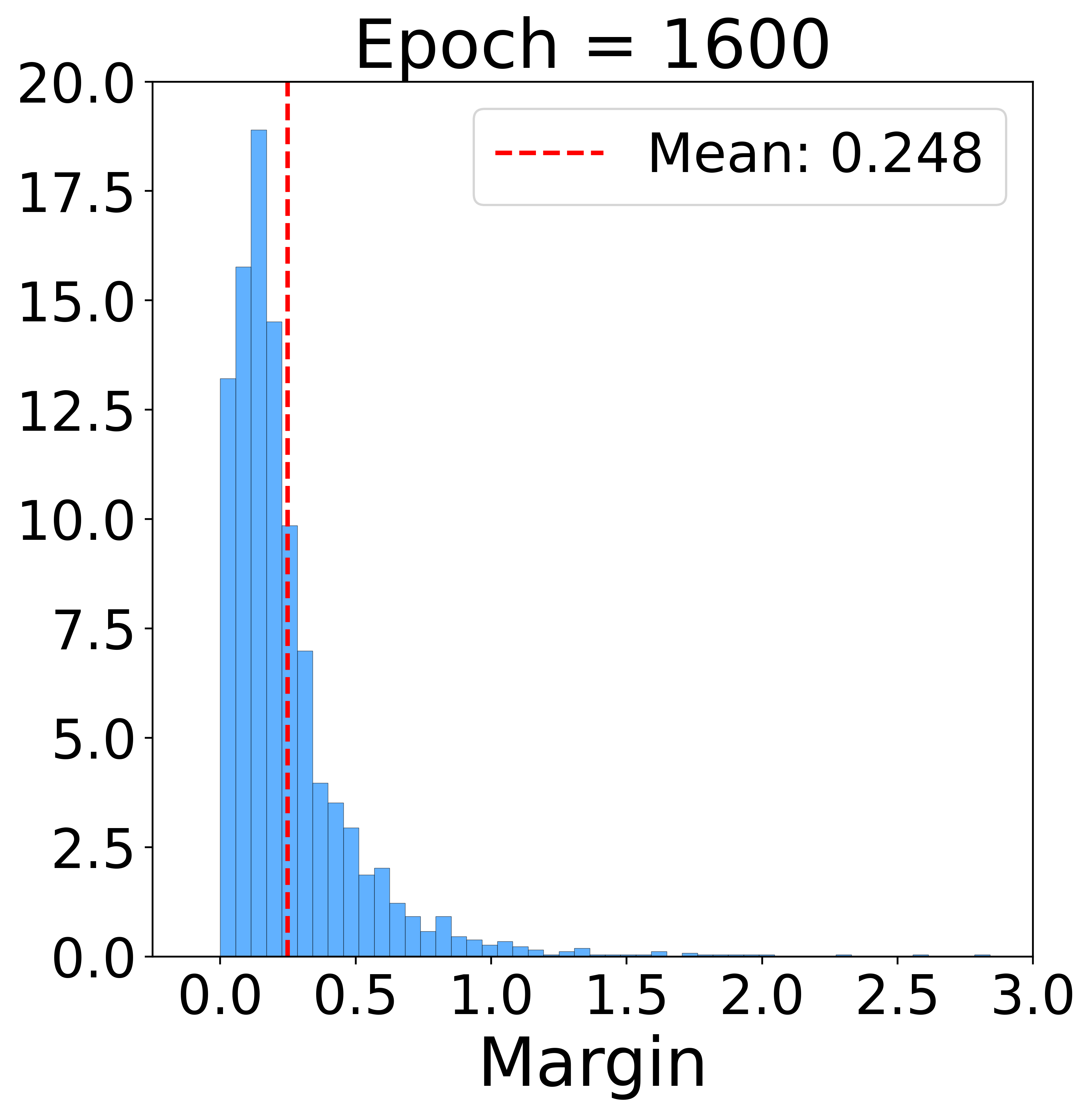}
    \end{subfigure}%
    \begin{subfigure}{0.19\linewidth}
        \centering
        \includegraphics[width=\linewidth]{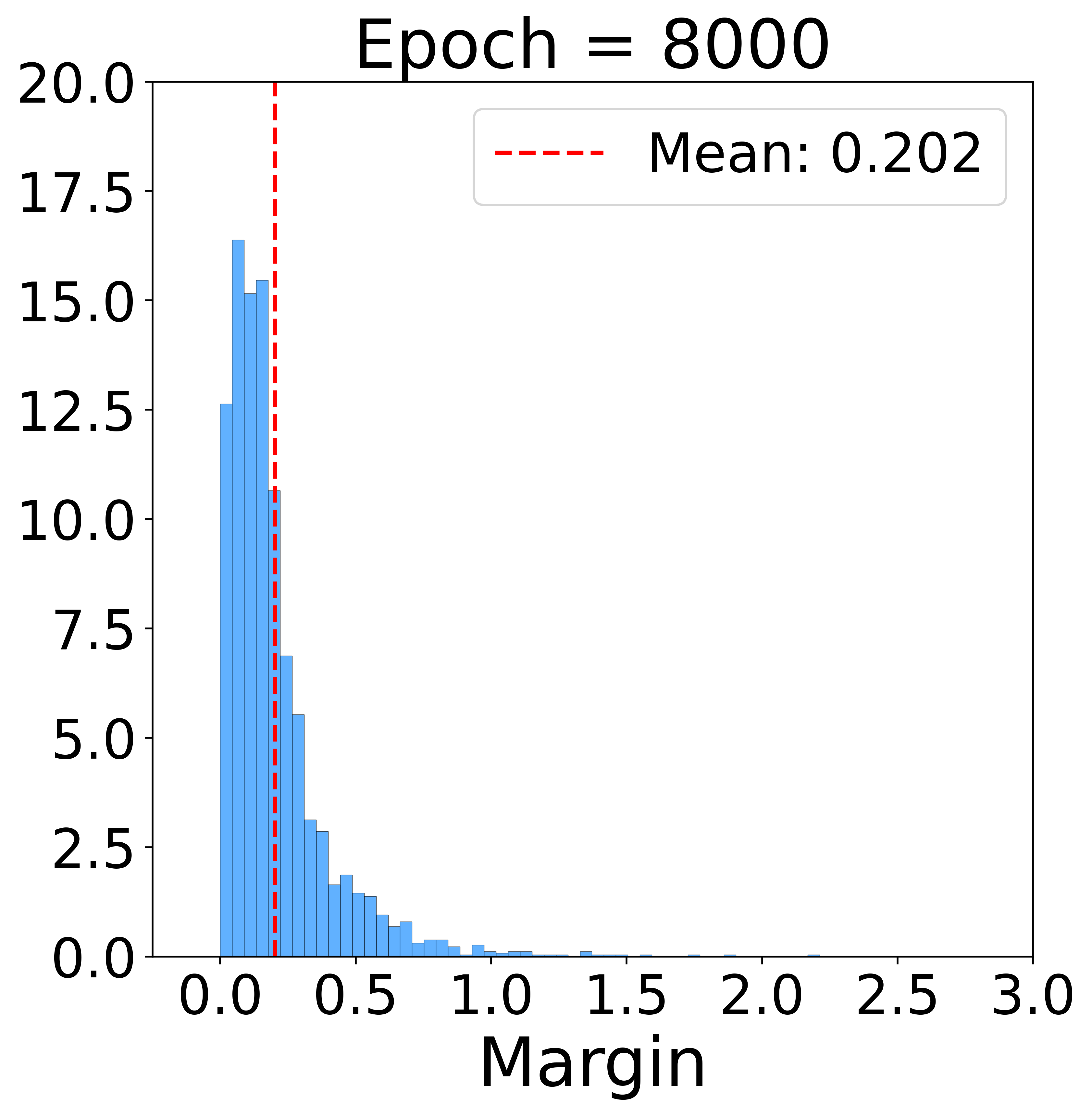}
    \end{subfigure}

    \caption{Evolution of the empirical distribution of margin distances for training data points in the \texttt{Water} dataset across different training epochs of a logistic regression model. As the training progresses, the \textit{average} margin distance decreases.}
    \label{fig:margin}
\end{figure*}

To further validate this claim, Figure~\ref{fig:margin} shows the empirical distribution of margin distances computed from all training points in the \texttt{Water} dataset~\cite{kadiwal2020waterpotability} at different training epochs of a logistic regression model.
We notice that, at initialization (epoch 0), the distribution is relatively uniform, with a higher average margin distance of $0.754$. As training progresses, the distribution of distances to the classifier's boundary becomes increasingly right-skewed, concentrating more of the mass near lower margin values. By epoch 8,000, the average margin distance drops to $0.202$.

\subsection{Estimating Counterfactual Probability}
\label{subsec:epsilon-VCF}
In this section, we formally characterize what we mean by the ease of finding an $\varepsilon$-valid counterfactual example for a given data point and, therefore, for a full training set of data points.

Let us consider the generic predictive model $\model$ trained on $\dataset = \{(\inst_i, y_i)\}_{i=1}^m$.
Suppose we associate to each training data point $\inst_i$ a binary random variable $X_i\in \{0,1\}$, which indicates whether there exists an $\varepsilon$-valid counterfactual example $\cfinst_i$ for $\inst_i$. 
Therefore, each $X_i$ follows a Bernoulli distribution, i.e., $X_i \sim \text{Bernoulli}(p_i^{\varepsilon})$, whose probability mass function is defined as below.
\begin{equation}
\Prob(X_i = k) = p_{X_i}(k;p_i^{\varepsilon}) = \begin{cases} p_i^{\varepsilon} &\text{, if }k=1
\\ 
1-p_i^{\varepsilon} &\text{, if } k=0.
\end{cases}
\end{equation}
Furthermore, let $\bar{X}$ denote the fraction of training points for which an $\varepsilon$-valid counterfactual example exists. Formally, $\bar{X}=\frac{1}{m}\sum_{i=1}^m X_i$ is the average of $m$ Bernoulli random variables. 

By the linearity of expectation, we can calculate:
\begin{equation}
\E[\bar{X}] = \E\Big[ \frac{1}{m}\sum_{i=1}^m X_i\Big] = \frac{1}{m}\sum_{i=1}^m \E[X_i],
\end{equation}
where $\E[X_i] = p_i^{\varepsilon}$ and, therefore, $\E[\bar{X}] = \bar{p}^{\varepsilon}$ is the average across the entire dataset.
Note that the random variable $Z=\sum_{i=1}^m X_i$ follows a Poisson-Binomial distribution, since $X_i$'s are independent but not necessarily identically distributed. Consequently, $\bar{X} = \frac{Z}{m}$ is also a Poisson-Binomial random variable, scaled by a factor of $1/m$.


For a fixed $\varepsilon > 0$, we argue that $\mathbb{E}[\bar{X}]$ is positively correlated with the model's training accuracy. 
This follows from the theoretical result discussed above \cite{wu2019arxiv}, which shows that higher training accuracy corresponds to a smaller average margin to the decision boundary. Therefore, as the training accuracy of $\model$ increases, so does the average fraction of training instances for which a valid counterfactual can be found within distance $\varepsilon$.

Firstly, we need to estimate each $p_i^{\varepsilon}$, which we refer to as the (sample-level) $\varepsilon$-\textit{valid counterfactual probability} ($\varepsilon$-VCP). 
Formally, let $\inst \in \X\subseteq \R^n$ be a generic training point, and $\varepsilon \in \R_{>0}$ a positive real number. Consider the $\varepsilon$ $n$-ball as the $n$-dimensional hypersphere of radius $\varepsilon$ centered around $\inst$. Let $V_{\inst}^{\varepsilon}\subseteq \R^n$ be the total hypervolume of this hypersphere, and let $V_{\cfinst}^{\varepsilon} \subseteq V_{\inst}^{\varepsilon}$ denote the portion of the total hypervolume falling within the counterfactual region delimited by the decision boundary induced by $\model$. Therefore, we can estimate the $\varepsilon$-VCP for $\inst$, as $p^{\varepsilon} = V_{\cfinst}^{\varepsilon}/V_{\inst}^{\varepsilon}$. 

\begin{figure}
 \centering
 \begin{subfigure}[htb!]{.38\linewidth}
 \includegraphics[width=.9\linewidth]{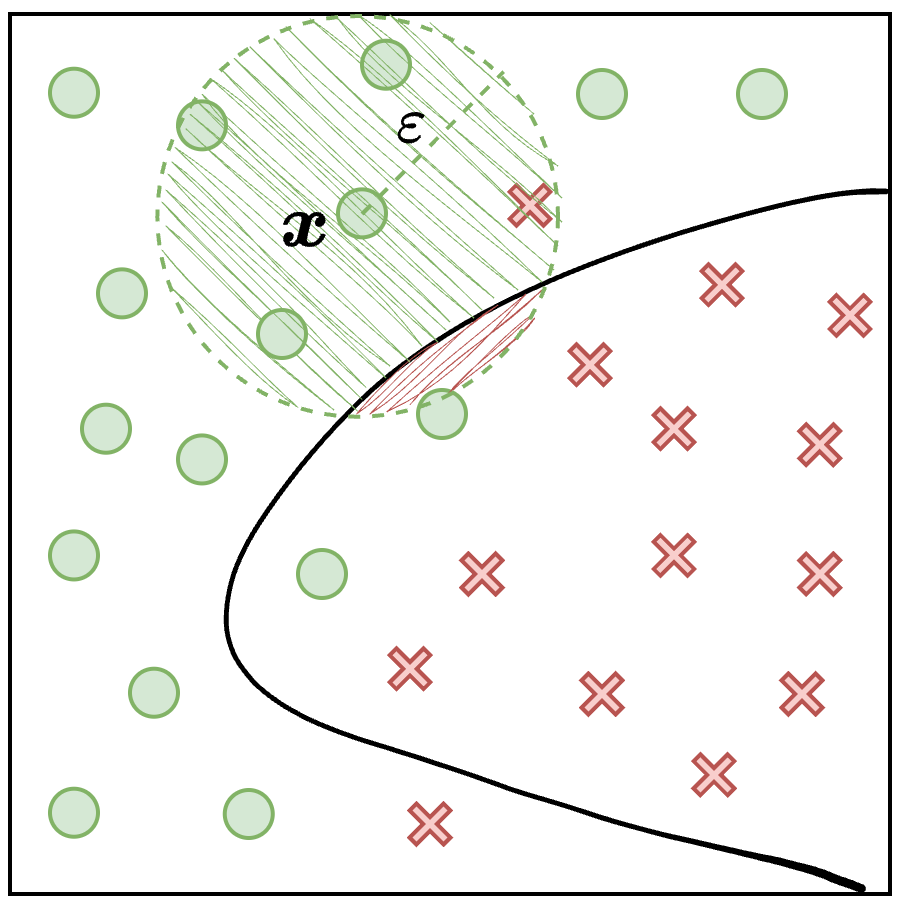}
 \caption{Low $\varepsilon$-valid counterfactual probability}
 \end{subfigure}
 \qquad
 \begin{subfigure}[h]{.38\linewidth}
 \includegraphics[width=.9\linewidth]{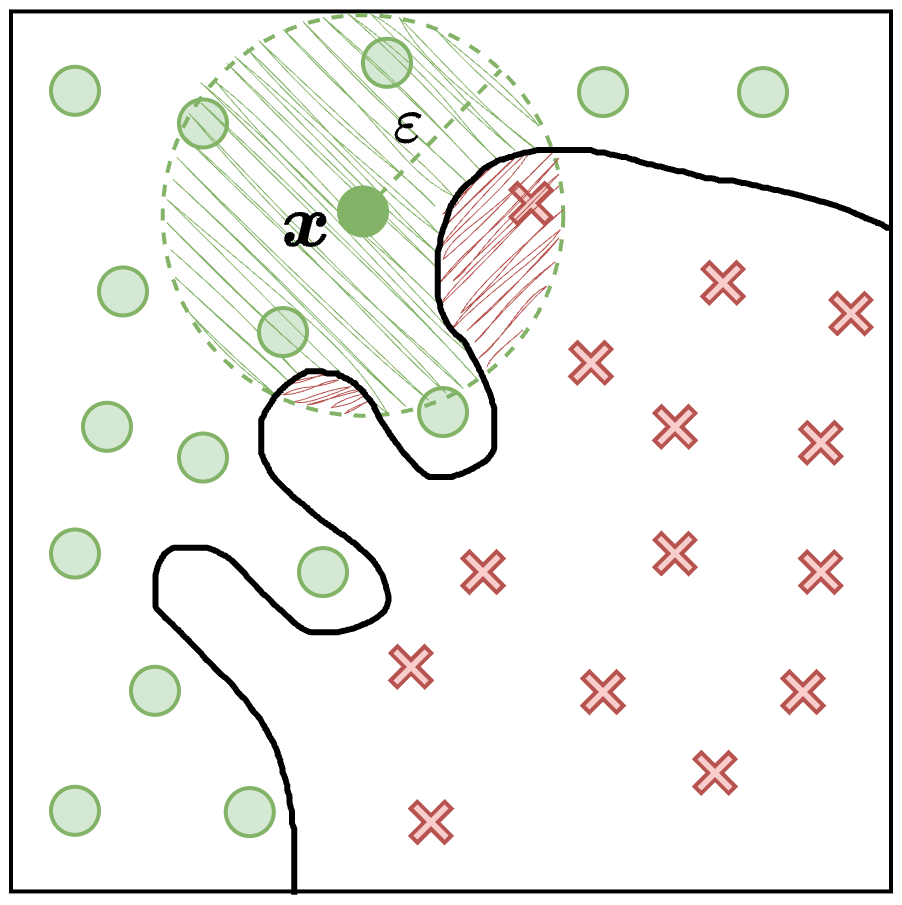}
 \caption{High $\varepsilon$-valid counterfactual probability}
 \end{subfigure}%
 \caption{The $\varepsilon$-\textit{valid counterfactual probability} for a sample $\inst \in \R^2$ can be estimated as the ratio of the area of the circle centered in $\inst$ with radius $\varepsilon$ that falls behind the decision boundary (in red).}
 \label{fig:cf-probability}
 \end{figure}

Intuitively, given a fixed $\inst$ and two models having different decision boundaries, it will generally be much easier to find an $\varepsilon$-valid counterfactual if the hypervolume $V_{\cfinst}^{\varepsilon}$ is larger.
This intuition is depicted in Figure~\ref{fig:cf-probability}, which illustrates how the probability of finding an $\varepsilon$-valid counterfactual for a $2$-dimensional data point may increase with model overfitting.
 
To estimate $V_{\cfinst}^{\varepsilon}$, we can apply standard Monte Carlo integration, a well-established technique that uses random draws to numerically compute a definite integral.
Furthermore, using Monte Carlo integration will provide valuable knowledge on the shape of a model's decision boundary as a by-product. 
The estimate outlined above assumes that data points are uniformly distributed around $\inst$ in the input space, which may be reasonable for appropriate values of $\varepsilon$. 
Indeed, it is worth noticing that $\inst$ can be an embedding from an autoencoding process, mapping each raw training point into a dense, lower-dimensional latent manifold within the original, high-dimensional space. 
More accurate estimates could be obtained by sampling the $\varepsilon$-neighborhood from the manifold using advanced strategies (e.g., see \citet{chen2022cikm}). However, this lies beyond the primary scope of this step and will be explored in future work.

Finally, if we extend the reasoning above to \textit{all} training points, we should observe that $\E[\bar{X}]=\bar{p}^{\varepsilon}$ is higher for overtrained models.

\subsection{Empirical Validation}
\label{subsec:empirical-validation}
To investigate the impact of overfitting on $\varepsilon$-VCP, we trained two Multi-Layer Perceptron (MLP) models on the \texttt{Water} dataset \cite{kadiwal2020waterpotability} for a binary classification task. Both models used the same 5-layer architecture, but only one employed dropout regularization with a rate of $0.5$. The models were trained over 500 epochs using the Adam optimizer with a learning rate of $\eta = 0.001$. The $\varepsilon$-VCP was estimated through Monte Carlo integration, as discussed in Section~\ref{subsec:epsilon-VCF}. Specifically, it was computed as $p^{\varepsilon} = V_{\cfinst}^{\varepsilon} / V_{\inst}^{\varepsilon}$, where $V_{\cfinst}^{\varepsilon}$ was approximated using 100 random samples per training point $\inst$.

It is important to note that the choice of $\varepsilon$ is inherently dependent on the characteristics of the dataset and is best determined empirically to ensure a meaningful evaluation of counterfactual validity while preserving consistency with the data distribution. As $\varepsilon$ defines the maximum norm of the perturbation vector applied to each instance, it delineates the feasible search space for generating counterfactuals.
An overly small value may excessively constrain the perturbations, limiting the ability to capture substantive changes in model predictions. In contrast, a value that is too large can produce counterfactuals that are implausible or inconsistent with the underlying data structure. 
In this experiment with the \texttt{Water} dataset, we found that setting $\varepsilon=1.5$ provides a suitable trade-off, allowing us to capture a diverse yet interpretable range of counterfactual examples without introducing excessive modifications to the original training points.

\begin{figure}
    \centering
    \includegraphics[width=0.8\linewidth]{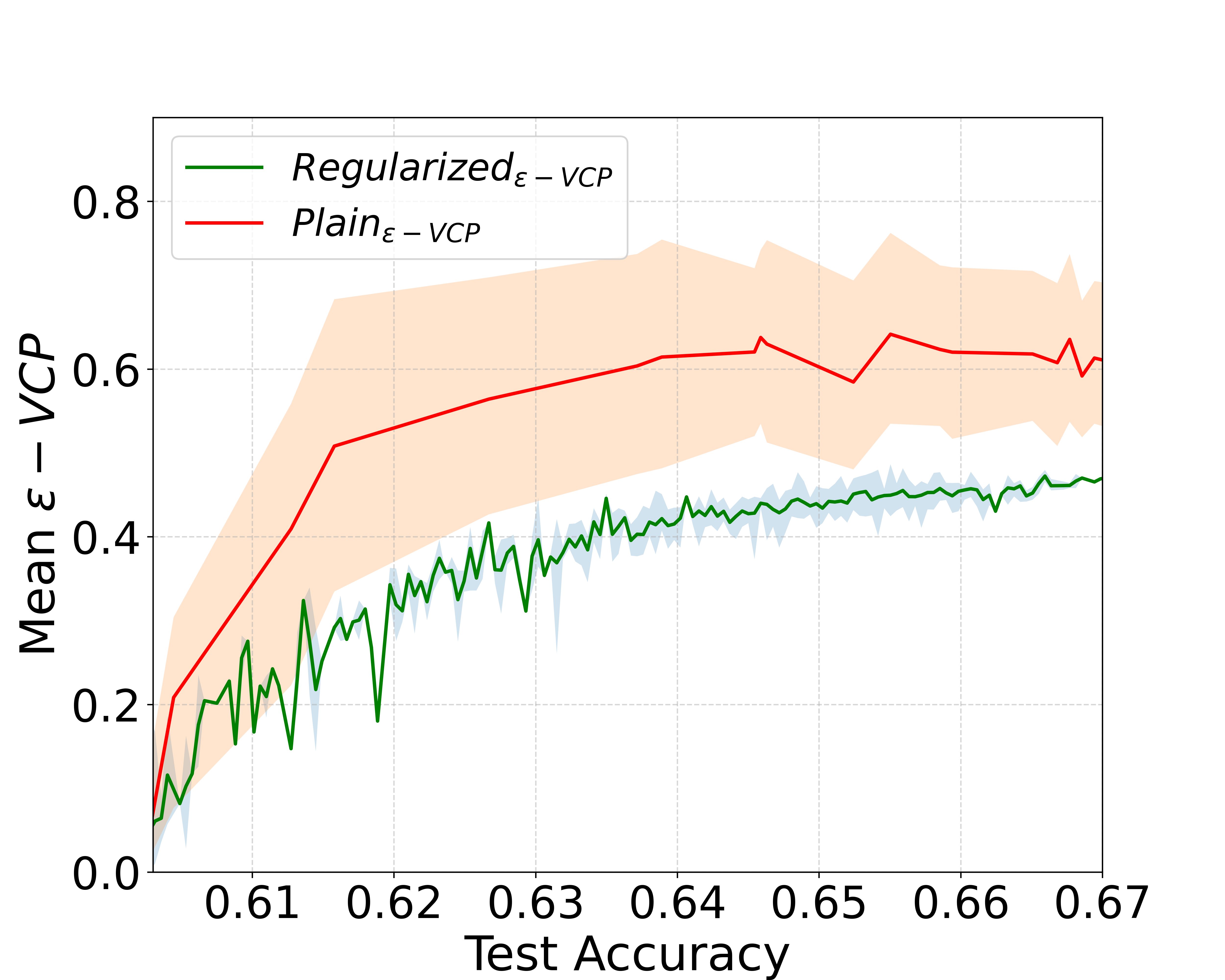}
    \caption{The mean $\varepsilon$-VCP ($y$-axis) vs. the model's training accuracy ($x$-axis). \textit{Plain}$_{\varepsilon\text{-VCP}}$ is the ``vanilla'' MLP, while \textit{Regularized}$_{\varepsilon\text{-VCP}}$ is the same MLP yet with a dropout rate of $0.5$.}
    \label{fig:evcp}
\end{figure}

Figure~\ref{fig:evcp} illustrates the evolution of the average $\varepsilon$-VCP, calculated over all training points, as a function of the models' training accuracy.
From this plot, three key insights emerge.
First, the average $\varepsilon$-VCP increases alongside training accuracy for both models, confirming our hypothesis that the likelihood of finding valid counterfactuals rises as models tend to overfit.
Second, the plain, unregularized MLP exhibits higher $\varepsilon$-VCP values, suggesting that its more complex decision boundary makes it easier to identify valid counterfactual examples.
Third, the trend persists while the MLP trained with dropout regularization exhibits a less pronounced increase (i.e., smaller $\varepsilon$-VCP values). This suggests that although dropout regularization smooths the decision boundary to some extent, it may not sufficiently counteract this phenomenon, leaving room for further improvement.

\section{Counterfactual Regularization (CF-Reg)}
\label{sec:method}
\subsection{A New Training Loss}
\label{subsec:new-loss}
We exploit the correlation between overfitting and the ability to find counterfactual examples, as highlighted in the previous section, to define a new regularized training loss as follows.
\begin{equation}\label{eq:regularized_loss}
\params^* = \argmin_{\params}\Big\{\Loss_{\text{emp}}(\params;\dataset) - \alpha \Loss_{\text{cf}}(\params;\dataset,\varepsilon)\Big\}.
\end{equation}
The first term ($\Loss_{\text{emp}}$) is the standard empirical risk, while the second term ($\Loss_{\text{cf}}$) is our proposed \textit{counterfactual regularization} component, with $\alpha \in \R_{\geq 0}$ serving as a hyperparameter to weigh its contribution. 
More specifically, the counterfactual regularization component can be defined as follows:
\begin{equation}\label{eq:cf_loss}
\Loss_{\text{cf}}(\params;\dataset,\varepsilon) = \varphi(\{d(\inst_i, \cfmodel(\inst_i)),\, \inst_i \in \mathcal{D} \};\varepsilon),
\end{equation}
where $\varphi (\cdot;\varepsilon)$ -- parameterized by $\varepsilon$ -- acts as an aggregation function applied to the set of distances $d(\inst_i, \cfmodel(\inst_i))$ between each sample $\inst_i$ and its corresponding counterfactual generated by the model, $\cfmodel(\inst_i)$.
Intuitively, the closer an instance is to its counterfactual, the greater the penalty imposed by the newly introduced loss. In other words, optimizing the objective defined in (\ref{eq:regularized_loss}) aims to ensure a sufficient margin between each training point and its corresponding counterfactual. 
Moreover, we would like this penalty to be stronger for training points where finding a counterfactual is easier -- i.e., those closer to the decision boundary.
To achieve this goal, each distance $d(\inst_i, \cfmodel(\inst_i))$ can be assigned a weight $w_i^{\varepsilon}$ that depends on $\varepsilon$ during aggregation. For example, this weight may be set as the $\varepsilon$-VCP associated with each training instance $\inst_i$ (i.e., $w_i^{\varepsilon} = p_i^{\varepsilon}$). 

Note that, for each data point $\inst_i$, its corresponding $p_i^{\varepsilon}$ can be estimated as described in Section~\ref{subsec:epsilon-VCF}. However, since $p_i^{\varepsilon}$ depends on the shape of the decision boundary, which may evolve dynamically across training epochs, its estimation should ideally be updated at \textit{every} epoch. Therefore, this could introduce a significant computational overhead in the training process, which we can mitigate updating our estimates $\{p_i^{\varepsilon}\}_{i=1}^m$ periodically, for example, every $t$ epochs.
We conjecture that a trade-off exists between the effectiveness of the counterfactual regularization term 
 -- impacted by the accuracy of each estimate $p_i^{\varepsilon}$ -- and the computational cost of the training process.


It is worth noticing that our counterfactual regularizer -- hereinafter referred to as CF-Reg -- is flexible enough to support any choice of aggregation function ($\varphi$), distance ($d$), and counterfactual generator ($\cfmodel$). However, the optimization problem in (\ref{eq:regularized_loss}) can be efficiently solved using standard gradient-based methods, provided that $\varphi$, $d$, and $\cfmodel$ are all differentiable with respect to $\params$.

In this work, we set $\varphi$ as the mean and $d$ as the Euclidean distance (i.e., the $L_2$-norm of the displacement vector resulting from the difference between the original sample and its counterfactual). 
For $\cfmodel$, we adopt the \textit{score counterfactual explanation} method proposed by~\citet{wachter2017hjlt}, and we discuss the rationale behind this choice below.
Alternative formulations of these components are possible and will be explored in future studies.

\subsection{Counterfactual Example Generator}
A key component of CF-Reg is the counterfactual generator $\cfmodel$, as this is used to compute the optimal counterfactual example for each training data point.
To incorporate this step into the regularized training process, the counterfactual generator must satisfy two essential requirements. First, it must be differentiable with respect to the predictive model weights $\params$. Second, it must be highly efficient to ensure the overall feasibility of our method. 

Among the various counterfactual generation approaches in the literature, we selected the \textit{score counterfactual explanation} method proposed by \cite{wachter2017hjlt}, which operates as follows.
Let $\inst \in \X$ be an input sample, $\model$ a trained model, $\delta:\X\times \X \mapsto \R_{\geq 0}$ a distance function, $\beta \in \R_{\geq 0}$ a controlling hyperparameter, and $s \in \R$ a target score. Thus, the score counterfactual explanation $\cfinst$ for $\inst$ is the result of the following optimization problem: 
\begin{equation}
\label{eq:sce}
        \cfinst^* = \underset{\cfinst \in \X}{\text{arg min}} \, (\model(\cfinst) - s)^2 + \beta \delta(\inst, \cfinst).
    \end{equation}

With a slight abuse of notation, in (\ref{eq:sce}), we treat the output of the predictive model $\model$ as a continuous value, even for classification tasks. 
To accommodate this requirement, we can, for instance, assume access to the logits, which are then passed through the appropriate activation function, such as sigmoid or softmax.

Note that the resulting counterfactual generator satisfies the two requirements mentioned above. In particular, by carefully selecting the distance function $\delta$ in (\ref{eq:sce}), the resulting objective becomes differentiable with respect to the predictive model weights 
$\params$. Furthermore, when using the score-based counterfactual explanation approach, determining the optimal counterfactual perturbation vector $\perturb = \inst - \cfinst^* = \inst - \cfmodel(\inst)$ -- i.e., the displacement vector between the original instance and its optimal counterfactual -- admits a closed-form solution under specific assumptions. Specifically, if $\model$ is a linear function and $\cfinst^* = \cfmodel(\inst)$ is the optimal counterfactual example for the input $\inst$, \citet{pawelczyk2022ijcai} proved that:
\begin{equation}
\label{eq:opt-delta}
    \perturb = \frac{t}{\beta + \|\params\|^2} \cdot \params,
\end{equation}
where $t = s - \model(\inst)$.
This makes the method highly efficient and well-suited for incorporation into our regularized training loss.

From (\ref{eq:opt-delta}), we can, therefore, compute the $L_2$-norm of the perturbation vector $\perturb_i$, for each training instance $\inst_i$. This value is then used as $d(\inst_i, \cfinst^*_i)$, as required by the proposed regularized training loss in (\ref{eq:cf_loss}).
Finally, the $L_2$-norms of all perturbation vectors are aggregated using the function $\varphi(\cdot;\varepsilon)$, resulting in the following (weighted) average: 
\begin{equation}
\label{eq:agg-delta}
\overline{||\perturb||} = \frac{1}{m}\sum_{i=1}^m w_i^{\varepsilon}*||\perturb_i||.
\end{equation}
In this work, weights are uniformly set as $w_i^{\varepsilon} = 1~\forall i\in \{1,\ldots,m\}$, thus reducing the aggregation to a plain average. Alternatively, more sophisticated strategies can be used, such as the one outlined in Section~\ref{subsec:new-loss}, where each training point is assigned a weight corresponding to its estimated $\varepsilon$-VCP, i.e., $w_i^{\varepsilon} = p_i^{\varepsilon}$.

Overall, this transforms our counterfactual regularized loss into: 
\begin{equation}\label{eq:regularized_loss2}
\params^* = \argmin_{\params}\Big\{\Loss_{\text{emp}}(\params;\dataset) - \alpha \overline{||{\perturb}||} \Big\}.
\end{equation}

Finally, the objective in (\ref{eq:regularized_loss2}) can be solved using standard gradient-based optimization methods.

\section{Experiments}
\label{sec:experiments}

In this section, we evaluate the ability of the proposed counterfactual regularized loss (CF-Reg), as defined in (\ref{eq:regularized_loss2}), to mitigate overfitting when compared to existing regularization techniques. 
In Appendix~\ref{app:avg-perturbation}, instead, we analyze the relationship between the test loss and the counterfactual vector norm during training.

\subsection{Experimental Setup}
\label{subsec:setup}
\textbf{\textit{Datasets, Models, and Tasks.}}
The experiments are run on four datasets: \texttt{Water}~\cite{kadiwal2020waterpotability}, \texttt{Phoneme}~\cite{phoneme}, \texttt{Higgs}~\cite{higgs}, and \texttt{CIFAR-10}~\cite{krizhevsky2009learning}. 
Furthermore, we consider the following models: Logistic Regression (LR), two Multi-Layer Perceptrons (MLP$_\text{small}$ and MLP$_\text{large}$), and \textit{PreactResNet-18}~\cite{he2016cvecvv} as a representative Convolutional Neural Network (CNN) architecture.
We focus on binary classification tasks and leave the extension to multiclass scenarios for future work. However, for datasets originally designed for multiclass classification, we adapt them by selecting two classes to align with our binary setting.
Additional details are provided in Appendix~\ref{app:setup}.

\smallskip
\noindent\textbf{\textit{Evaluation Measures.}} To characterize the degree of overfitting, we use the test loss, as it serves as a reliable indicator of the model's generalization capability to unseen data. Additionally, we evaluate the predictive performance of each model using the test accuracy.

\smallskip
\noindent\textbf{\textit{Baselines.}} We compare CF-Reg with the following methods: Early Stopping, Dropout, adversarial training via well-known projected gradient descent (PGD)~\cite{madry2018pgd}, two regularizers -- i.e., L1-Reg (``Lasso'') and L2-Reg (``Ridge'') -- and no regularization (No-Reg).

\smallskip
\noindent \textbf{\textit{Hyperparameter Settings.}}  
We analyze the sensitivity of CF-Reg to its hyperparameters in Section~\ref{subsec:hyperparameters}. Due to space constraints, a comprehensive discussion covering all regularization methods, including the tuning strategies adopted, is provided in Appendix~\ref{app:hyperparameters}. The best selected values are summarized in Table~\ref{tab:performance_parameters}.

\smallskip
\noindent \textbf{\textit{Implementation Details.}} To deliberately induce overfitting in logistic regression models, we apply a polynomial feature expansion that increases the input dimensionality beyond the number of training samples. This ensures that the model has sufficient capacity to memorize the training data, thereby allowing us to assess the effect of our counterfactual regularizer. The degree of the polynomial expansion is selected as the smallest value that results in a number of features exceeding the number of training instances. Note that this preprocessing step is not required for neural network models, as they are chosen to be complex enough to overfit the data.

In addition, to leverage the closed-form solution for computing the optimal perturbation vector as defined in \eqref{eq:opt-delta}, we use a local linear approximation of the model during the counterfactual generation process. Thus, given an instance $\inst_i$, we compute the (optimal) counterfactual not with respect to the original model $\model$, but with respect to its first-order Taylor expansion around $\inst_i$, denoted as:
\begin{equation}
\label{eq:taylor}
    \model^{lin}(\inst) = \model(\inst_i) + \nabla_{\inst}\model(\inst_i)(\inst - \inst_i).
\end{equation}
Note that this step is unnecessary for logistic regression, as it is inherently a linear model.

We run all experiments on a machine equipped with an AMD Ryzen 9 7900 12-Core Processor and an NVIDIA GeForce RTX 4090 GPU. Our implementation is based on the PyTorch Lightning framework, and the source code for our experiments is available at the following GitHub repository: \url{https://anonymous.4open.science/r/CF-Reg-5AB6/README.md} 

Further details are described in Appendix \ref{app:training}.

\subsection{Effectiveness of CF-Reg}
We compare the performance of CF-Reg against the baselines indicated in Section~\ref{subsec:setup}. 
For each model and dataset combination, Table~\ref{tab:regularization_comparison} shows the mean value and standard deviation of test accuracy achieved by each method across 5 random initializations. 
From the results in Table~\ref{tab:regularization_comparison}, the following key insights can be drawn. 
\smallskip

\noindent \textbf{\textit{General Validity of CF-Reg.}}
Across the board, CF-Reg demonstrates strong performance, particularly on tabular datasets such as \texttt{Water}, \texttt{Phoneme}, and \texttt{Higgs}. 
In many configurations, it either matches or outperforms all other baselines, and in several cases, the improvements are statistically significant. This supports the core hypothesis behind CF-Reg: that training with informative counterfactual examples can improve generalization by encouraging decision boundary robustness in data-relevant directions.
\smallskip

\noindent \textbf{\textit{Comparison with other Regularization Techniques.}}
CF-Reg consistently outperforms traditional penalty-based regularizers like L1-Reg and L2-Reg. These methods penalize model complexity indirectly through weight constraints, whereas CF-Reg directly encourages decision-level stability, which appears more effective in preserving model generalization.
Moreover, CF-Reg performs on par with or better than implicit regularization methods such as early stopping and dropout. Notably, in cases where CF-Reg underperforms slightly, it still retains a critical advantage: it leverages the full expressive capacity of the architecture, rather than curtailing it through early termination of training or by randomly omitting connections during forward passes. This is particularly relevant for modern architectures where expressiveness is key to capturing subtle data patterns.
When compared to adversarial training (PGD), CF-Reg also shows almost always superior or comparable accuracy.

\smallskip

\noindent \textbf{\textit{Handling Complex Input Domains.}}
On the \texttt{CIFAR-10} dataset, CF-Reg does not outperform traditional regularizers. In fact, in this context, the No-Reg baseline performs the best, suggesting that overfitting is less of a concern, perhaps due to the indirect regularizing effect of convolutional architectures. 
Additionally, the current counterfactual generation strategy appears less effective for image data, where generating meaningful perturbations is substantially harder. This points to a promising direction for future work: using more expressive and domain-aware counterfactual generators tailored to high-dimensional, unstructured inputs, in order to fully extend the applicability of CF-Reg to complex data.

\begin{table*}[h!]
    \centering
    \caption{Mean and standard deviation of test accuracy over 5 random initializations for each combination of model, dataset, and regularization method. Best results are shown in bold; results that are statistically significantly better than the second-best (at the $\alpha = 0.05$ significance level) are marked with a $^{\star}$.}
    \label{tab:regularization_comparison}
    \resizebox{\textwidth}{!}{%
    \begin{tabular}{|c|c|c|c|c|c|c|c|c|}
        \hline
        \textbf{Model} & 
        \textbf{Dataset} & 
        \textbf{No-Reg} & 
        \textbf{L1-Reg} & 
        \textbf{L2-Reg} & 
        \textbf{Dropout} & 
        \textbf{Early Stopping} & 
        \textbf{PGD} & 
        \textbf{CF-Reg (ours)} \\ \hline
        
        LR   & \texttt{Water}   & $0.6030 \pm 0.0053$   & $0.6652 \pm 0.0053$   & $0.6677 \pm 0.0114$  & N/A    & $0.6098 \pm 0.0000$  & $0.6384\pm0.0183$ & $\mathbf{0.6915 \pm 0.0017^\star}$                   \\ \hline
        MLP$_\text{small}$   & \texttt{Water}   & $0.6128 \pm 0.0103$   & $0.6098 \pm 0.0000$   & $0.6759 \pm 0.0065$  & $0.6515 \pm 0.0100$    & $ 0.6765 \pm 0.0059$    & $0.6524\pm0.0011$& $\mathbf{0.6796 \pm 0.0045}$                \\ \hline
        MLP$_\text{large}$   & \texttt{Water}   & $0.6168 \pm 0.0063$   & $0.6098 \pm 0.0000$   & $0.6320 \pm 0.0158$  & $0.6564 \pm 0.0104$    & $ 0.6573 \pm 0.0112$    & $0.6464\pm0.0053$ & $\mathbf{0.6787 \pm 0.0092^{\star}}$                \\ \hline
        LR   & \texttt{Phoneme}   & $0.8729 \pm 0.0052$   & $0.8157 \pm 0.0036$   & $0.8427 \pm 0.0078$  & N/A    & $0.8622 \pm 0.0109$       & $0.7353\pm0.0006$ & $\mathbf{0.8764 \pm 0.0084}$              \\ \hline
        MLP$_\text{small}$   & \texttt{Phoneme}   & $\mathbf{0.9016 \pm 0.0088}$   & $0.8511 \pm 0.0092$   & $0.7963 \pm 0.0041$  & $\mathbf{0.9016 \pm 0.0059}$ & $0.8957 \pm 0.0105$ & $0.8915\pm0.0029$ & $0.9005 \pm 0.0065$                   \\ \hline
        MLP$_\text{large}$   & \texttt{Phoneme}   & $0.9101 \pm 0.0026$   & $0.7068 \pm 0.0000$   & $0.8735 \pm 0.0042$  & $0.9099 \pm 0.0078$     & $0.9066 \pm 0.0083$ & $0.9021\pm0.0021$ & $\mathbf{0.9149 \pm 0.0024^{\star}}$                   \\ \hline
        MLP$_\text{small}$   & \texttt{Higgs}   & $0.7204 \pm 0.0043$   & $0.4706 \pm 0.0000$   & $0.7231 \pm 0.0033$  & $0.7210 \pm 0.0016$     & $0.7271 \pm 0.0049$ & $\boldsymbol{0.7334\pm0.0025^{\star}}$ & $0.7245 \pm 0.0057$                   \\ \hline
        MLP$_\text{large}$   & \texttt{Higgs}   & $0.6904 \pm 0.0020$   & $0.4706 \pm 0.0000$   & $0.7223 \pm 0.0056$  & $0.7323 \pm 0.0039$     & $0.7149 \pm 0.0043$ & $0.6995\pm0.0018$ & $\mathbf{0.7278 \pm 0.0027}$                   \\ \hline
        CNN   & 
        \texttt{CIFAR-10} & 
        $\mathbf{0.8415 \pm 0.0059}$   & 
        $0.8355 \pm 0.0160$   & 
        $0.8408 \pm 0.0035$   & 
        N/A    &  
        $0.8413 \pm 0.0008$ &  
        $0.7071 \pm 0.0316$ &  
        $0.8388 \pm 0.0106$ \\ \hline 
    \end{tabular}
    }
    \end{table*}

\begin{table*}[htb!]
    \centering
    \caption{Mean and standard deviation of training time (in seconds) over 5 independent runs. Each value refers to the training process for the corresponding entry in Table~\ref{tab:regularization_comparison}.}
    \label{tab:times}
    \resizebox{\textwidth}{!}{
    \begin{tabular}{|c|c|c|c|c|c|c|c|c|}
        \hline
        \textbf{Model} & 
        \textbf{Dataset} & 
        \textbf{No-Reg} & 
        \textbf{L1-Reg} & 
        \textbf{L2-Reg} & 
        \textbf{Dropout} &
        \textbf{Early Stopping} &
        \textbf{PGD} & 
        \textbf{CF-Reg (ours)} \\ \hline
        
        LR   & 
        \texttt{Water}   & 
        $19.86 \pm 0.51$   & 
        $20.87 \pm 0.33$   & 
        $21.06 \pm 0.29$  & 
        N/A    & 
        $0.46 \pm 0.11$ &
        $119.23\pm0.83$ &
        $21.78 \pm 0.54$                     \\ \hline
        
        MLP$_\text{small}$   & 
        \texttt{Water}   & 
        $19.78 \pm 0.21$   & 
        $22.47 \pm 0.47$   & 
        $22.60 \pm 0.41$  & 
        $21.20 \pm 0.26$    & 
        $1.37 \pm 0.49$    &
        $120.31\pm0.05$ &
        $23.07 \pm 0.28$                    \\ \hline
        
        MLP$_\text{large}$   & 
        \texttt{Water}   & 
        $22.94 \pm 0.37$   & 
        $26.25 \pm 0.18$   & 
        $27.21 \pm 0.42$   & 
        $25.88 \pm 0.45$    &
        $0.76 \pm 0.10$    &
        $157.83\pm0.05$ &
        $27.48 \pm 0.29$                    \\ \hline
        
        LR   & 
        \texttt{Phoneme}   & 
        $29.54 \pm 0.40$ & 
        $32.25 \pm 0.68$   & 
        $32.22 \pm 0.23$  & 
        N/A   & 
        $20.40 \pm 3.27$  &
        $128.67\pm 0.05$ &
        $33.84 \pm 0.37$                    \\ \hline
        
        MLP$_\text{small}$   &
        \texttt{Phoneme} & 
        $31.02 \pm 0.56$   & 
        $34.90 \pm 0.35$   & 
        $35.55 \pm 0.72$   & 
        $33.12 \pm 0.18$   &
        $31.33 \pm 0.07$  &
        $565.60\pm1.67$ &
        $36.60 \pm 0.72$                     \\ \hline
        
        MLP$_\text{large}$   &
        \texttt{Phoneme} & 
        $35.66 \pm 0.60$   & 
        $41.27 \pm 0.99$   & 
        $42.60 \pm 0.59$ & 
        $40.25 \pm 0.70$    &
        $20.95 \pm 0.70$  &
        $253.80\pm0.83$ &
        $43.95 \pm 0.56$                     \\ \hline
        
        MLP$_\text{small}$   &
        \texttt{Higgs} & 
        $415.16 \pm 4.70$   & 
        $470.94 \pm 8.46$   & 
        $485.45 \pm 8.22$   & 
        $448.73 \pm 5.04$    &
        $156.89 \pm 7.03$  &
        $459.40\pm1.67$ &
        $506.11 \pm 5.73$                     \\ \hline
        
        MLP$_\text{large}$   &
        \texttt{Higgs} & 
        $486.08 \pm 5.35$   & 
        $571.48 \pm 10.70$   &
        $593.64 \pm 8.38$ & 
        $545.84 \pm 23.19$    &
        $29.38 \pm 2.73$  &
        $584.10\pm1.41$  &
        $613.39 \pm 10.71$                     \\ \hline
        
        CNN   & 
        \texttt{CIFAR-10} & 
        $377.47 \pm 1.66$   &
        $401.37 \pm 1.94$   & 
        $404.53 \pm 3.54$ & 
        N/A    & 
        $318.19 \pm 2.20$  &
        $1304.40\pm5.62$ &
        $1294.23 \pm 3.88$                     \\ \hline
    \end{tabular}
    }
\end{table*}

\subsection{Hyperparameter Sensitivity Analysis}
\label{subsec:hyperparameters}
CF-Reg relies on two key hyperparameters: $\alpha$ and $\beta$. The former is intrinsic to the loss formulation defined in (\ref{eq:cf-train}), while the latter is closely tied to the choice of the score-based counterfactual explanation method used.

\begin{figure}[ht]
    \centering
    \includegraphics[width=0.85\linewidth]{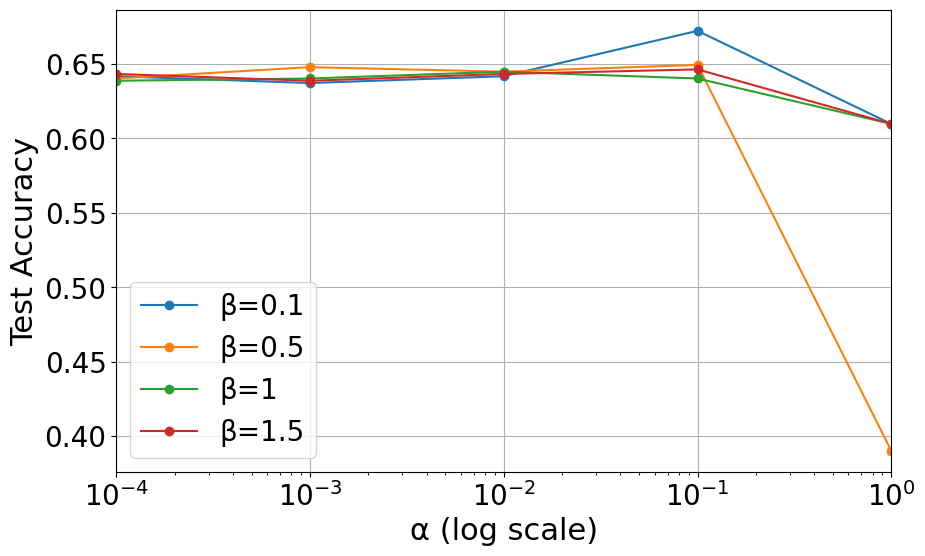}
    \caption{The test accuracy of an MLP trained on the \texttt{Water} dataset, evaluated while varying the weight of our counterfactual regularizer ($\alpha$) for different values of $\beta$.}
    \label{fig:test_alpha_beta}
\end{figure}

Figure~\ref{fig:test_alpha_beta} illustrates how the test accuracy of an MLP trained on the \texttt{Water} dataset changes for different combinations of $\alpha$ and $\beta$.

We observe that, for a fixed $\beta$, increasing the weight of our counterfactual regularizer ($\alpha$) can slightly improve test accuracy until a sharp drop occurs for $\alpha > 0.1$.
This behavior is expected, as the impact of CF-Reg, like any regularization term, can be disruptive if not properly calibrated.
Moreover, this finding confirms that CF-Reg is inherently data-driven and, thus, requires specific fine-tuning based on the combination of the model and dataset at hand.

The best values for $\alpha$ and $\beta$ are shown in Table~\ref{tab:performance_parameters} (Appendix \ref{app:hyperparameters}).

\section{Discussion}
\label{sec:discussion}

\subsection{Feasibility of CF-Reg}
\label{subsec:feasibility}
As with any regularizer, CF-Reg introduces some computational overhead in exchange for improved model generalization. Specifically, the extra cost stems from the need to compute a counterfactual example \textit{for each} training instance during learning. Since CF-Reg is compatible with any counterfactual generator, the actual overhead depends on the complexity of the generator used.

Indeed, more advanced counterfactual generators may yield more effective regularization but require longer computation times. Conversely, simpler generators, such as the score-based model employed in this study, offer efficient training at the potential cost of regularization quality. We analyze this trade-off in greater depth in Section~\ref{subsec:limitations}.

We quantify the cost of CF-Reg by evaluating its impact on model training time. Specifically, Table~\ref{tab:times} reports the mean and standard deviation of training time for each model-dataset pair presented in Table~\ref{tab:regularization_comparison}. 
We observe that CF-Reg introduces only a modest computational overhead when employing the score-based generator. However, on larger datasets, the need to compute counterfactual distances for every training point leads to increased training time, as expected. Interestingly, the number of model parameters does not significantly affect the training time. Instead, input dimensionality plays a more substantial role, especially for nonlinear models. For example, in \texttt{CIFAR-10}, gradients for local linearization must be computed in a high-dimensional space, $\mathbb{R}^{3 \times 32 \times 32}$, which contributes considerably to the increased training cost.

\subsection{Counterfactual Explanations as a By-Product}
\label{subsec:explanations}
While CF-Reg introduces some computational overhead and does not always significantly outperform all baselines, it offers a unique advantage not provided by traditional regularizers: the generation of counterfactual explanations as a natural by-product of training. 
These explanations, typically computed through separate and often costly post hoc procedures~\cite{chen2022cikm,movin2025tai}, are instead amortized over training with CF-Reg, yielding improved generalization, possibly without additional inference cost.

At test time, given a new input $\inst_{\text{new}}$, approximate counterfactual explanations for the prediction $\model(\inst_{\text{new}})$ can be efficiently retrieved by: $(i)$ identifying the $k$ nearest training instances to $\inst_{\text{new}}$, and $(ii)$ returning their precomputed counterfactual examples via a simple lookup, which takes constant time ($O(1)$) per instance.

\subsection{Main Limitations and Future Improvements}
\label{subsec:limitations}
In this section, we discuss the main limitations of CF-Reg and outline directions for future improvements. Our method was designed to be general and efficient; however, some design choices, while practical, may limit performance in certain settings. 
Below, we sketch the key areas that are worth investigating.

\smallskip
\noindent \textbf{\textit{Efficiency vs. Regularization Trade-off.}}  
As introduced in Section~\ref{subsec:new-loss}, our framework supports \textit{any} counterfactual generator $\cfmodel$, but this flexibility introduces variability in computational cost. When using lightweight generators (e.g., score-based), training remains fast, but the regularization may be less effective. In contrast, more expressive generators could improve generalization at the expense of longer training times. 
Exploring this trade-off, and the use of more sophisticated counterfactual generators, is a promising avenue for future research.

To improve the feasibility of CF-Reg, we plan to investigate the following strategies: $(i)$ Cache and update counterfactuals every $t$ epochs instead of at each iteration; $(ii)$ Cluster training points and generate one counterfactual per cluster rather than per sample; $(iii)$ Perform counterfactual search only for a selected subset of influential training points, identified via random sampling or heuristic criteria.

\smallskip
\noindent \textbf{\textit{Linear Approximation.}}  
Even when using the score-based generator, we do not fully leverage the closed-form solution for computing the optimal perturbation vector in (\ref{eq:opt-delta}), which assumes a linear model $\model$. For deep neural networks, this assumption generally does not hold. We instead use the first-order Taylor approximation in (\ref{eq:taylor}), which may yield suboptimal counterfactuals due to its inability to fully capture the model's nonlinear behavior.

\smallskip
\noindent \textbf{\textit{Uniform Weighting Scheme.}}  
Our current implementation assigns equal importance to each training instance ($w_i^{\varepsilon} = 1$). While simple, this uniform weighting may limit the potential of CF-Reg. A more fine-grained weighting strategy could better balance the regularization strength across samples, though it would likely increase training complexity. Investigating adaptive weighting mechanisms remains an open and worthwhile direction.

\smallskip
\noindent \textbf{\textit{Counterfactual Generator Learning.}}
In this work, we treat the counterfactual generator $\cfmodel$ as a fixed component, queried during the regularized training of the primary model $\model$.
A promising direction for future research is to integrate the learning of $\cfmodel$ directly into the training process, thereby formulating the task as a bilevel optimization problem.
Solving this problem would allow for the joint learning of both the predictive model $\model$ and its associated counterfactual generator $\cfmodel$.

\section{Conclusion}
\label{sec:conclusion}
We introduced and analyzed the trade-off between the generalizability and explainability of predictive models, highlighting their often conflicting nature. 
Specifically, based on well-known theoretical results in margin theory, we demonstrated that the degree of model overfitting is positively correlated with its ability to generate counterfactual examples. 
Building on this insight, we proposed CF-Reg, a novel regularization technique that integrates a counterfactual regularization term into the training objective, to balance between predictive performance and interpretability.

Through extensive experiments across multiple datasets and model architectures, we showed that CF-Reg generally outperforms existing regularization techniques, improving generalization while simultaneously generating meaningful counterfactual explanations. Our results suggest that counterfactual-based regularization can serve as a principled approach to improving model robustness without sacrificing interpretability.

Finally, we outlined promising directions for future work, primarily aimed at improving the efficiency of our method.

\bibliographystyle{ACM-Reference-Format}
\bibliography{references}

\newpage
\clearpage
\appendix

\section{Additional Experimental Details}
\label{app:experiments}

\subsection{Datasets, Models, and Tasks}
\label{app:setup}
Three of the four datasets used in this study, namely \texttt{Water}, \texttt{Phoneme}, and \texttt{Higgs}, are tabular, while the fourth, \texttt{CIFAR-10}, is a popular image dataset.
Table~\ref{tab:datasets} summarizes key characteristics of each dataset, including the number of samples, number of features, class distribution (minority and majority class sizes), and a brief description.

\begin{table*}[!bt]
\caption{Main properties and description of the datasets.}
\centering
\renewcommand{\arraystretch}{1.2}
  \begin{tabularx}{\textwidth}{@{}L{12mm}L{12mm}L{12mm}L{22mm}L{22mm}X@{}}
    \toprule
    \textbf{Dataset} & \textbf{\#Samples} & \textbf{\#Features} & \textbf{\#Min.~Class (\%)} & \textbf{\#Maj.~Class (\%)} & \textbf{Description} \\ 
    \midrule
    \texttt{Water}   & 3,276      & 9          & 1,278 ($39\%$)      & 1,998 ($61\%$)      & This tabular dataset includes various water quality metrics and is used to predict water potability, i.e., whether it is safe for human consumption. \\
    \texttt{Phoneme} & 5,404      & 5          & 1,586 ($29\%$)      & 3,818 ($71\%)$      & This tabular dataset contains five attributes selected to characterize each vowel, with the goal of distinguishing between nasal and oral sounds. \\
    \texttt{Higgs}   & 200,000    & 28         & 94,131 ($47\%$)      & 105,869 ($53\%$)    & This dataset is a random sample without replacement of the original \texttt{Higgs} tabular dataset, comprising $18\%$ of the total 1.1 million instances, and is used to distinguish between a signal process that produces Higgs bosons and a background process that does not.\\
    \texttt{CIFAR-10} & 12,000     & 3×32×32    & 6,000 ($50\%$)       & 6,000 ($50\%$)      & This dataset is a sample of the original \texttt{CIFAR-10} image dataset, adapted to a binary classification setting, containing only instances from class 3 (``cats'') and class 5 (``dogs''). \\
    \bottomrule
  \end{tabularx}%
\label{tab:datasets}
\end{table*}
\FloatBarrier
All datasets except \texttt{CIFAR-10} are naturally associated with a binary classification task.
To adapt \texttt{CIFAR-10} to a binary classification setting, we retain only instances from class 3 (``cats'') and class 5 (``dogs'').

For the \texttt{Water}, \texttt{Phoneme}, and \texttt{Higgs} datasets, we performed an 80/20 train-test split. For \texttt{CIFAR-10}, we used the standard train-test split provided by the \texttt{torchvision} library.

\FloatBarrier
Table~\ref{tab:model-comparison} summarizes the key characteristics of the models evaluated in our experiments. 
The selection covers a broad spectrum of architectures, from simple logistic regression models with a few thousand parameters to deep neural networks comprising tens of millions of parameters.
\begin{table}[!htb]
\caption{Comparison of various models across datasets, layer configurations, and parameter counts.}
  \centering
  \begin{tabularx}{\columnwidth}{llcc}
    \toprule
    \textbf{Model} & \textbf{Dataset} & \textbf{Layers Width} & \textbf{\#Parameters} \\
    \midrule
    LR                   & \texttt{Water}      & N/A                         & 5,005    \\
    MLP$_\text{small}$   & \texttt{Water}      & [100, 30]                   & 3,930    \\
    MLP$_\text{large}$     & \texttt{Water}      & [150, 1000, 150, 30]        & 305,880 \\
    \midrule
    LR                   & \texttt{Phoneme}    & N/A                         & 4,368   \\
    MLP$_\text{small}$   & \texttt{Phoneme}    & [100, 40]                   & 4,540   \\
    MLP$_\text{large}$     & \texttt{Phoneme}    & [150, 1000, 150, 30]        & 305,280 \\
    \midrule
    MLP$_\text{small}$   & \texttt{Higgs}      & [100, 30]                   & 5,830   \\
    MLP$_\text{large}$     & \texttt{Higgs}      & [150, 1000, 150, 30]        & 308,730 \\
    \midrule
    CNN                  & \texttt{CIFAR-10}   & ResNet-18                   & 11,200,000 \\
    \bottomrule
  \end{tabularx}
  \label{tab:model-comparison}
\end{table}
\FloatBarrier

\subsection{Training Settings}
\label{app:training}
Table~\ref{tab:optimizers} illustrates the training settings used for each dataset. 
We adopt Adam as optimizer due to its faster convergence properties. This choice enables us to reduce the number of training epochs while still performing an exhaustive hyperparameter search within the same computational budget. 

\begin{figure}[!htb]
    \centering
    \includegraphics[width=\columnwidth]{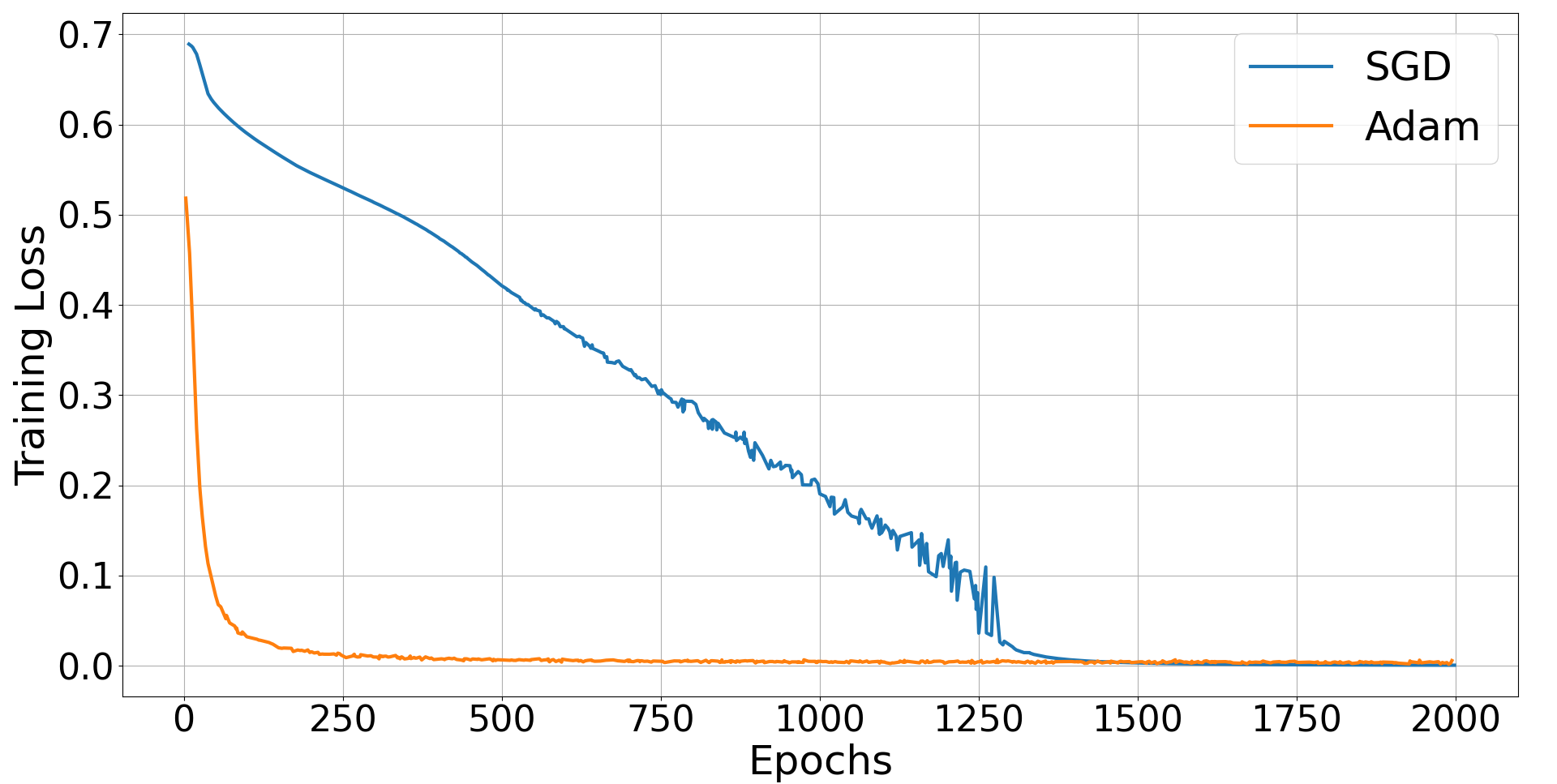}
    \caption{The training loss evolution over training epochs for an MLP trained on the \texttt{Higgs} dataset without regularization with SGD (blue) or Adam (orange) as optimizer. Adam shows a faster fit to training data in terms of training epochs.}
    \label{fig:optimizer_sgd_vs_adam}
\end{figure}

To illustrate Adam's faster convergence in practice, Figure~\ref{fig:optimizer_sgd_vs_adam} compares its performance to standard stochastic gradient descent (SGD) when training an MLP on the \texttt{Higgs} dataset.
\FloatBarrier
\begin{table}[!htb]
\caption{Training settings for each dataset.}
\label{tab:optimizers}
\centering
\scalebox{0.88}{
\begin{tabular}{lcccc}
  \toprule
  \textbf{Dataset} & \textbf{Optimizer} & \textbf{\#Epochs} & \textbf{Batch Size} & \textbf{Learning Rate} ($\eta$)\\
  \midrule
  \texttt{Water}   & Adam & 2,000 & 128 & $0.001$ \\
  \texttt{Phoneme} & Adam & 2,000 & 128 &  $0.001$ \\
  \texttt{Higgs}   & Adam & 200 & 128 & $0.001$   \\
  \texttt{CIFAR-10} & Adam & 200 & 128 & $0.001$   \\
  \bottomrule
\end{tabular}
}
\end{table}
\FloatBarrier
\begin{table*}[htb!]
    \centering
    \caption{Hyperparameter settings used to generate the results in Table~\ref{tab:regularization_comparison}. L1-Reg and L2-Reg report the regularization coefficients, i.e., the weights of the $L_1$ and $L_2$ norms of the model parameters, respectively. Dropout indicates the dropout probability. Early Stopping shows the patience parameter. PGD reports: the step size of the adversarial attack ($\alpha$); the maximum perturbation budget ($\epsilon$); and the number of attack iterations ($k$). Lastly, CF-Reg represents the regularization coefficients $\alpha$ and $\beta$.}
    \label{tab:performance_parameters}
    \begin{tabular}{|c|c|c|c|c|c|c|c|}
        \hline
        \textbf{Model} & 
        \textbf{Dataset} & 
        \textbf{L1-Reg} & 
        \textbf{L2-Reg} & 
        \textbf{Dropout} & 
        \textbf{Early Stopping} & 
        \textbf{PGD} & 
        \textbf{CF-Reg (ours)} \\ \hline
        
        LR   & 
        \texttt{Water}   & 
        $4.933\,e\text{-}02$   &
        $7.413\,e\text{-}01$  & 
        N/A    & 
        $5$    &
        $1.076\,e\text{-}02/1.128\,e\text{-}02/15$ & 
        $3.353\,e\text{-}01/9.816\,e\text{-}01$
        \\ \hline
        
        MLP$_\text{small}$   & 
        \texttt{Water}   & 
        $1.280\,e\text{-}02$    & 
        $2.920\,e\text{-}03$    & 
        $2.004\,e\text{-}01$    & 
        $15$    &
        $1.364\,e\text{-}01/4.714\,e\text{-}02/05$ &
        
        $8.325\,e\text{-}01/1.886\,e\text{+}00$                    \\
        \hline
        
        MLP$_\text{large}$   & 
        \texttt{Water}   & 
        $1.605\,e\text{-}02$    & 
        $2.613\,e\text{-}03$    & 
        $8.192\,e\text{-}01$    & 
        $10$    &
        $9.430\,e\text{-}01/1.614\,e\text{-}02/05$ &

        $1.121\,e\text{+}00/2.118\,e\text{+}00$                    \\
        \hline
        
        LR   &
        \texttt{Phoneme}   &
        $1.796\,e\text{-}02$   & 
        $3.659\,e\text{-}02$  & 
        N/A    & 
        $260$    &
        $1.543\,e\text{-}02/6.009\,e\text{-}03/05$ &
        $1.303\,e\text{-}05/3.461\,e\text{-}02$                     \\
        \hline
        
        MLP$_\text{small}$   & 
        \texttt{Phoneme}   & 
        $1.097\,e\text{-}03$   & 
        $6.972\,e\text{-}03$  & 
        $1.606\,e\text{-}01$    & 
        $280$    &
        $1.100\,e\text{-}02/9.069\,e\text{-}02/20$ &

        $1.485\,e\text{-}02/1.883\,e\text{-}01$                    \\ \hline
        
        MLP$_\text{large}$   &
        \texttt{Phoneme}   & 
        $5.885\,e\text{-}03$   &
        $3.003\,e\text{-}03$  & 
        $1.001\,e\text{-}02$    & 
        $260$    &
        $1.044\,e\text{-}02/7.132\,e\text{-}03/05$ &

        $5.539\,e\text{-}03/1.797\,e\text{-}01$                    \\ \hline
        
        MLP$_\text{small}$   & 
        \texttt{Higgs}   & 
        $6.667\,e\text{-}03$   &
        $3.214\,e\text{-}04$  &
        $9.820\,e\text{-}02$    & 
        $50$    &
        $3.099\,e\text{-}02/2.153\,e\text{-}02/05$ &
        $1.097\,e\text{-}01/4.724\,e\text{-}01$                    \\ \hline
        
        MLP$_\text{large}$   & 
        \texttt{Higgs}   & 
        $1.391\,e\text{-}02$    & 
        $1.675\,e\text{-}04$    & 
        $3.918\,e\text{-}01$    & 
        $5$    &
        $1.422\,e\text{-}02/8.403\,e\text{-}02/05$ &
        $4.380\,e\text{-}01/2.289\,e\text{+}00$\\ \hline
        
        CNN   & 
        \texttt{CIFAR-10} & 
        $8.700\,e\text{-}06$   & 
        $4.868\,e\text{-}06$ & 
        N/A    & 
        $160$    &
        $3.755\,e\text{-}01/5.043\,e\text{-}03/07$ &
        $7.246\,e\text{-}04/1.852\,e\text{-}01$\\ \hline
    \end{tabular}
\end{table*}
Training and testing are performed over 2,000 epochs for the \texttt{Water} and \texttt{Phoneme} datasets, and over 200 epochs for both \texttt{CIFAR-10} and \texttt{Higgs}.
In all cases, we use a batch size of 128, and we set the learning rate to $\eta = 0.001$ with no weight decay.

\section{Counterfactual Perturbation vs. Overfitting}
\label{app:avg-perturbation}
In Section~\ref{subsec:empirical-validation}, we empirically demonstrated that the average $\varepsilon$-VCP is correlated with the risk of overfitting: specifically, higher training accuracy tends to correspond to a larger average $\varepsilon$-VCP. That analysis provided a general observation, with the average $\varepsilon$-VCP estimated using Monte Carlo integration.

In this section, instead, we conduct a more targeted analysis by examining the relationship between the test loss (i.e., the complement of test accuracy) and the average $L_2$-norm of the counterfactual perturbation vectors ($\overline{||\perturb||}$) over the course of training. The latter serves as a practical proxy for the average $\varepsilon$-VCP.

\begin{figure}[!htb]
    \centering
    \includegraphics[width=0.85\linewidth]{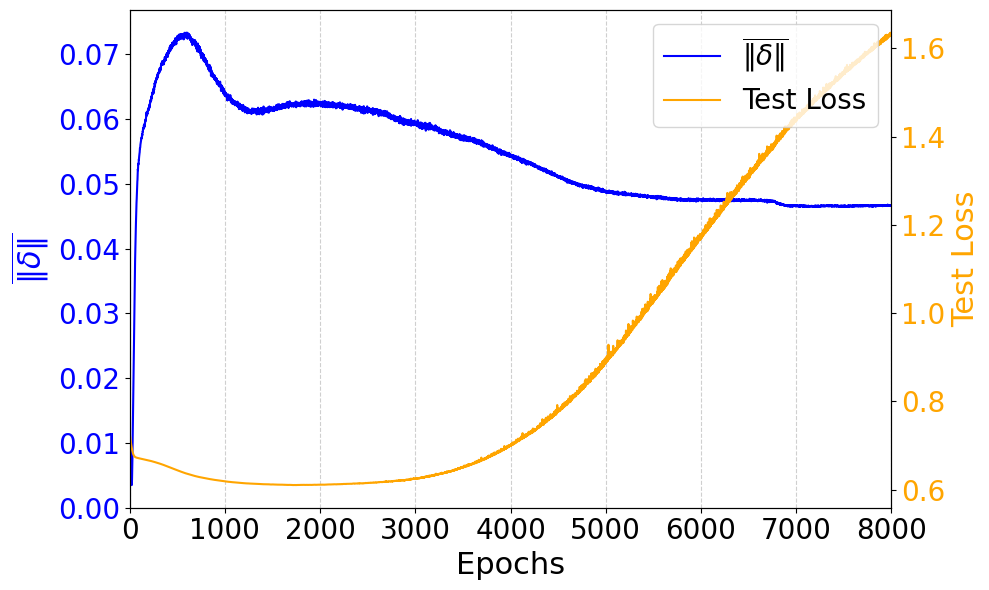}
    \caption{The average counterfactual perturbation vector $\overline{||\perturb||}$ (left $y$-axis) and the cross-entropy test loss (right $y$-axis) over training epochs ($x$-axis) for an MLP trained on the \texttt{Water} dataset \textit{without} regularization.}
    \label{fig:delta_loss_epochs}
\end{figure}

Figure~\ref{fig:delta_loss_epochs} illustrates the evolution of $\overline{||\perturb||}$ and the test loss over training epochs for an MLP trained without regularization on the \texttt{Water} dataset.
The plot shows a clear trend as the model starts to overfit the data (evidenced by an increase in test loss). 
Notably, $\overline{||\perturb||}$ begins to decrease, which aligns with the hypothesis that the average distance to the optimal counterfactual example gets smaller as the model's decision boundary becomes increasingly adherent to the training data.

It is worth noticing that this trend is heavily influenced by the choice of the counterfactual generator model. In particular, the relationship between $\overline{||\perturb||}$ and the degree of overfitting may become even more pronounced when leveraging more accurate counterfactual generators. However, these models often come at the cost of higher computational complexity, and their exploration is left to future work.

Nonetheless, we expect that $\overline{||\perturb||}$ will eventually stabilize at a plateau, as the average $L_2$-norm of the optimal counterfactual perturbations cannot vanish to zero.

\section{Hyperparameter Tuning} \label{app:hyperparameters}
Most regularization techniques are highly sensitive to hyperparameter choices.
Table~\ref{tab:performance_parameters} summarizes all hyperparameters used in this study. Each value was selected via random Bayesian optimization over approximately 80 trials, aiming to maximize accuracy on a held-out validation set.

In the table, L1-Reg and L2-Reg refer to the coefficients of the respective regularization terms, i.e., the weights of the $L_1$ and $L_2$ norms of the model parameters. The Dropout entry indicates the dropout probability applied within the model architecture. For Early Stopping, the reported value corresponds to the patience parameter, namely the number of epochs without improvement before training is stopped. The three entries listed under PGD correspond to: the step size of the adversarial attack ($\alpha$); the maximum perturbation budget ($\epsilon$); and the number of attack iterations ($k$). Finally, the two CF-Reg values represent the regularization coefficients $\alpha$ and $\beta$, as defined in \eqref{eq:regularized_loss} and \eqref{eq:sce}, respectively.

\end{document}